\documentclass[sn-mathphys-num]{sn-jnl}

\usepackage{graphicx}%
\usepackage{multirow}%
\usepackage{amsmath,amssymb,amsfonts}%
\usepackage{amsthm}%
\usepackage{mathrsfs}%
\usepackage[title]{appendix}%
\usepackage{xcolor}%
\usepackage{textcomp}%
\usepackage{subfig}%
\usepackage{makecell}%
\usepackage{multirow}
\usepackage{geometry}
\usepackage{multicol}
\usepackage{manyfoot}%
\usepackage{orcidlink}
\usepackage[linesnumbered,ruled]{algorithm2e}
\usepackage{algpseudocode}
\usepackage{booktabs}
\usepackage{listings}%
\usepackage{caption}
\usepackage{lscape}
\usepackage{adjustbox}
\theoremstyle{thmstyleone}%
\theoremstyle{thmstyletwo}%
\definecolor{cvprblue}{rgb}{0.21,0.49,0.74}
\DeclareMathOperator*{\argmax}{arg\,max}
\DeclareMathOperator*{\argmin}{arg\,min}
\newcommand{\mycomment}[1]{}

\theoremstyle{thmstylethree}%

\begin{document}

\title[Realistic Image-to-Image Machine Unlearning]{Realistic Image-to-Image Machine Unlearning via \\ Decoupling and Knowledge Retention}

\author*[]{\fnm{Ayush K.} \sur{Varshney}\orcidlink{0000-0002-8073-6784}}\email{ayushkv@cs.umu.se}
\author[]{\fnm{Vicen\c{c}} \sur{Torra}\orcidlink{0000-0002-0368-8037}}\email{vtorra@cs.umu.se}
\affil[]{\centering \orgdiv{Department of Computing Science}, \orgname{Umeå University}, \\ \orgaddress{\city{Umeå}, \postcode{907 36}, \country{Sweden}}}

\abstract{Machine Unlearning allows participants to remove their data from a trained machine learning model in order to preserve their privacy, and security. However, the machine unlearning literature for generative models is rather limited. The literature for image-to-image generative model (I2I model) considers minimizing the loss between Gaussian noise and the output of I2I model for forget samples as machine unlearning. However, we argue that the machine learning model performs fairly well on unseen data i.e., a retrained model will be able to catch generic patterns in the data and hence will not generate an output which is just Gaussian noise instead. In this paper, we consider that the model after unlearning should treat forget samples as out-of-distribution (OOD) data, i.e., the unlearned model should no longer recognize or encode the specific patterns found in the forget samples. To achieve this, we propose a framework which decouples the model parameters with gradient ascent, ensuring that forget samples are OOD for unlearned model with theoretical guarantee. We also provide $(\epsilon, \delta)$-unlearning guarantee for model updates with gradient ascent. The unlearned model is further fine-tuned on the remaining samples to maintain its performance. We also propose a data poisoning attack model as an auditing mechanism in order to make sure that the unlearned model has effectively removed the influence of forget samples. Furthermore, we demonstrate that even under sample unlearning, our approach prevents backdoor regeneration, validating its effectiveness. Extensive empirical evaluation on two large-scale datasets, ImageNet-1K and Places365 highlights the superiority of our approach. To show comparable performance with a retrained model, we also show the comparison of a simple AutoEncoder on various baselines on CIFAR-10 dataset. Code will be available after acceptance.}

\maketitle

\section{Introduction}

Generative machine learning, such as image processing ~\cite{dhariwal2021diffusion} and natural language processing~\cite{achiam2023gpt}, has made tremendous strides in recent years, driven by the availability of large-scale datasets and increased computational power. This allows them to learn complex patterns from the vast amount of available data and produce high-quality realistic outputs across various domains. However, these datasets often contain sensitive information. To safeguard the user privacy and information, regulations across the world allow users the right to remove their data and its influence from any machine learning (ML) model. 

Machine \textit{Un}learning~\cite{cao2015towards} has emerged as a paradigm that allows machine learning models to remove the influence of specific data samples (forget samples) without retraining the model, which can be computationally expensive. The goal is to remove the influence of forget samples while preserving the performance on the remaining samples (retain samples). Beyond the regulatory requirement, machine unlearning is also useful for removing the adversarial participants, addressing copyright infringement issues, and robustness of ML models. The literature of machine unlearning, such as a sharded training in~\cite{bourtoule2021machine} and modified fine-tuning in~\cite{tarun2023fast}, has primarily focused on the classification problem in machine learning. However, unlearning in generative ML has not received much attention.

Generative models are known to retain information from the training data~\cite{somepalli2023diffusion}, which can be exploited to regenerate original training samples~\cite{carlini2023extracting}.  This raises significant privacy concerns, making effective unlearning mechanisms essential. In this paper, we address the unlearning challenge of the specific architecture of generative models called Image-to-Image (I2I) generative models. Prior research on I2I machine unlearning~\cite{li2024machine} frames unlearning as an optimization problem where the objective is to minimize the distance between Gaussian noise and forget samples while fine-tuning on the retain samples to preserve performance. An extension of this was proposed in~\cite{feng2024controllable}, where a set of pareto-optimal solution is presented to address varying user expectations. However, ML models are expected to perform well on unseen data, in the context of unlearning, the retrained model from retain samples would still be able to predict generalized patterns on forget samples. This implies that unlearning is \textit{not equivalent} to merely replacing the influence of forget samples with Gaussian noise, as generic patterns may persist. We argue that for effective unlearning, the forget samples should be treated as out-of-distribution (OOD) samples for the unlearned model. OOD samples are those samples whose distribution significantly deviates from that of the training data. Current I2I unlearning literature~\cite{li2024machine, feng2024controllable} primarily focus on class-level or dataset-level unlearning, and to the best of our knowledge, there is no prior work on sample-level unlearning in I2I generative models. Furthermore, existing approaches lack formal unlearning guarantees.

\begin{figure*}
  \centering
  \begin{adjustbox}{width=1.2\textwidth,center}
    \includegraphics[]{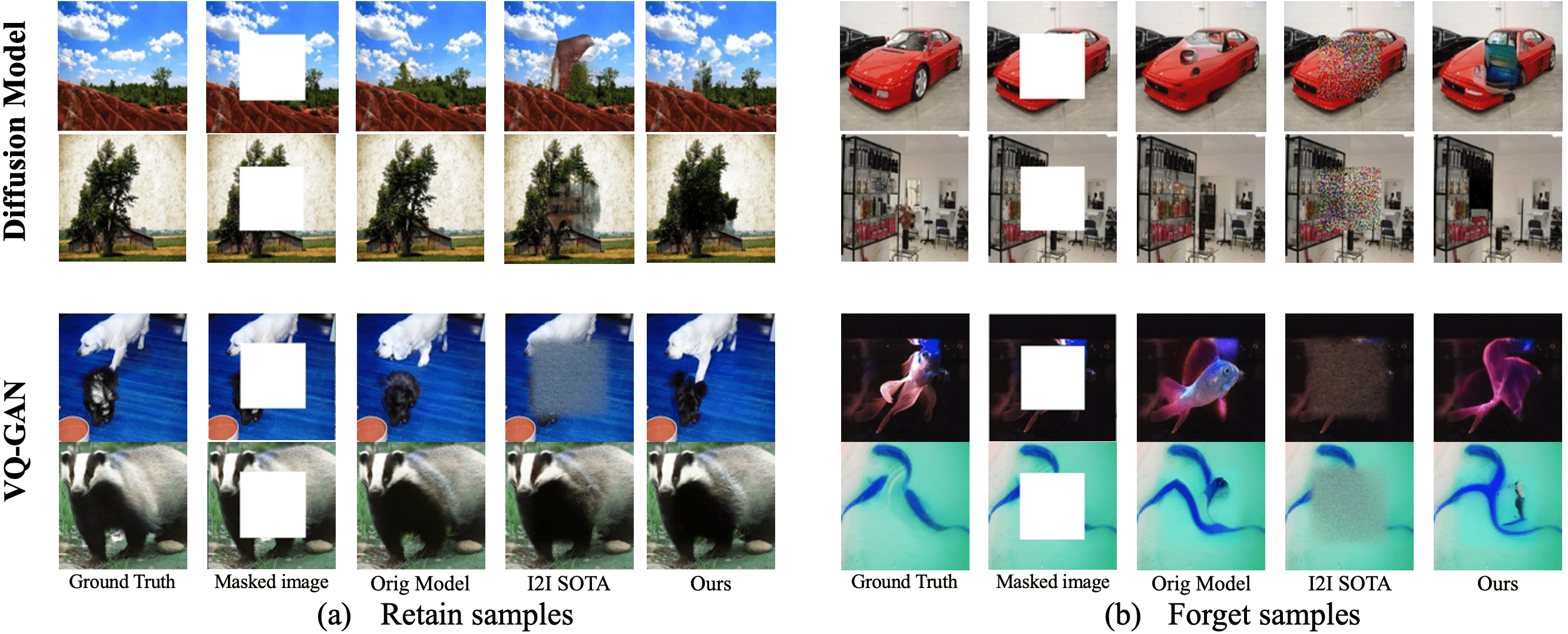}
    \end{adjustbox}
    \caption{Our approach is effective across major Image-to-Image (I2I) architectures, including VQ-GAN~\cite{li2023mage}, diffusion model~\cite{saharia2022palette}, and autoencoders (see Section 5). The figure also presents a comparison with the state-of-the-art (SOTA) I2I unlearning algorithm. For retain samples, our method generates consistent images before and after unlearning, while the SOTA method may generate inconsistent output. On forget samples, our approach intentionally produces inaccurate or unreliable outputs, aligning with the expected behavior of realistic unlearning.}
    \label{fig:Intro_Realistic_I2I}
\end{figure*}

To overcome these drawbacks, we propose a two-step unlearning approach. First, we decouple the model updates on forget samples with gradient ascent~\cite{halimi2022federated}. Gradient ascent (GA) maximizes the loss on forget samples. We show that, after performing $T$ iterations of GA, the model treats the forget samples as out-of-distribution data. Furthermore, we formally prove that the model updates after $T$ iterations of gradient ascent satisfy formal $(\epsilon, \delta)$-unlearning guarantees. In the second step, the model is further fine-tuned on retain samples in order to preserve its performance on them. This two-step process allows us to achieve realistic unlearning of the forget samples while maintaining the performance on the retain dataset. In the case of sample unlearning, the first step causes the forget samples to become out-of-distribution (OOD) samples. However, since similar patterns are present in the retain set, fine-tuning on retain samples can reintroduce these patterns, potentially making the forget samples appear in-distribution again. Despite this, the unlearning guarantees established after the first step remain valid for the final model. As shown in Fig. \ref{fig:Intro_Realistic_I2I}, our approach demonstrates consistent output quality in the context of image-inpainting, maintaining fidelity on retain samples while producing inconsistent output on forget samples to reflect effective unlearning. 



Auditing the effectiveness of unlearning is also critical. However, relying solely on performance metrics is insufficient, as noisy equivalents are likely to exhibit poor performance on the forget samples. To address this, we propose an alternative auditing mechanism based on a data poisoning attack on the forget samples. Specifically, we fine-tune the model on poisoned versions of the forget samples to embed a distinctive pattern during inference. After the unlearning process, an effective unlearning mechanism should prevent the unlearned model from reproducing this pattern, thereby confirming the successful removal of the poisoned data’s influence. 

In summary, we make the following contributions.
\begin{enumerate}
    \item A realistic image-to-image machine unlearning framework which treats forget samples as out-of-distribution data after unlearning. Our framework is backed by theoretical guarantees, demonstrating that forget samples are indeed classified as OOD and unlearning process satisfies $(\epsilon, \delta)$-unlearning guarantees.
    \item We propose a data poisoning attack to audit the effectiveness of the unlearning process.
    \item We introduce sample-level unlearning for I2I generative models, addressing a gap in existing literature.  
    \item An extensive evaluation of our algorithm on various I2I generative models, including autoencoders, VQ-GAN and diffusion model. The empirical results on two large scale dataset highlight the superiority of our framework.
    \item Additionally, to benchmark our approach against a fully retrained model, we provide a comparative analysis on the CIFAR-10 dataset, showcasing the effectiveness of our unlearning framework. 
\end{enumerate} 

The rest of the paper is organized as follows. Section \ref{sec:related work} briefly describes the related work on Image-to-Image (I2I) generative models and machine unlearning. Section \ref{sec:prelims} outlines the preliminaries our proposed framework.  Section \ref{sec:proposed_work} provides the details of our framework along with the theoretical unlearning guarantees. Section \ref{sec:exp_details} presents the experimental analysis. The paper finishes with some conclusions and future work in Section \ref{sec:conclusion}.

\section{Related Work}
\label{sec:related work}

\subsection{Image-to-Image Generative Model}

Many computer vision tasks can be formulated as Image-to-Image generative tasks ranging from de-colorization \cite{saharia2022palette}, image super-resolution \cite{bulat2018learn}, image-inpainting \cite{krishnan2019boundless}, image-outpainting \cite{chang2022maskgit}, and many more. The three main types of architectures used for these tasks are: autoEncoders \cite{alain2014regularized} (AEs), generative adversarial networks (GANs) \cite{goodfellow2020generative}, and diffusion models \cite{ho2020denoising}. AEs aim to minimize the mean squared error between the generated output and the ground truth, which works well for lower-quality outputs but struggles to capture fine details. In contrast, GANs generate realistic images by employing an adversarial framework, where a generator network attempts to create convincing fake images while a discriminator network tries to distinguish real from fake. Despite their effectiveness, GANs are notorious for unstable training and mode collapse. Diffusion models, on the other hand, generate high-quality outputs by gradually learning to denoise data after adding noise in successive steps. However, they require substantial amounts of data and computational resources, making them less efficient in resource-constrained environments. In this work, we aim to design generic unlearning framework for all the I2I models. 

\subsection{Machine Unlearning}

Machine Unlearning \cite{cao2015towards} is the process of efficiently removing the influence of specific data points (forget samples) from a trained machine learning model without the need to fully retrain the model from scratch. Machine unlearning approaches should also ensure that removing specific data does not adversely affect the performance of the model on the remaining data. The objective of machine unlearning is to produce a model that closely approximates the one we would get after retraining from scratch on the remaining samples. Given that retraining a model can be computationally expensive, machine unlearning offers a more efficient alternative, especially in cases where compliance with data privacy laws is necessary.


The concept of machine unlearning was initially introduced in the context of statistical query learning \cite{cao2015towards}. The introduction of SISA framework \cite{bourtoule2021machine} advanced this area by enabling efficient unlearning through selective retraining of the specific model checkpoints. However, retraining can be computationally expensive if retraining samples are distributed across shards. To mitigate this, several approaches have been proposed, such as estimating the influence of the data samples using inverse of the hessian matrix \cite{golatkar2020eternal}, leveraging Newton’s method \cite{guo2019certified}, or using the Fisher Information Matrix. Although these methods have shown promise, their application to large-scale datasets remains computationally prohibitive due to the high cost of matrix computations. Other approaches forgets by maximizing their loss on forget samples or fine-tuning the model on misclassified labels \cite{tarun2023deep, chen2023boundary}. However, these approaches have been used for classification tasks in general. 

Despite the progress in machine unlearning for classification tasks, the field has received limited attention in the context of generative machine learning \cite{liu2024machine} due to its inability to compare with retrained model. Due to which many existing studies focus on small datasets, such as unlearning experiments on MNIST \cite{bae2023gradient}, or feature unlearning \cite{moon2023feature}. Recently, a large-scale unlearning framework for image-to-image generative models was given in \cite{li2024machine}, which considers minimizing the distance between the generated output for forget samples and Gaussian noise as an unlearning objective. A variation of this \cite{feng2024controllable} considers $\epsilon$-constrained unlearning objective, where $\epsilon$ controls the degree of unlearning. However, we argue that a more realistic unlearning approach should ensure that the forget samples be treated as out-of-distribution data by the unlearned model, as would be the case with a model fully retrained without those samples. Hence, this motivates us to explore the realistic unlearning algorithms for image-to-image generative models in large-scale setup.


\section{Preliminaries}
\label{sec:prelims}



\subsection{Out-Of-Distribution data}

Out-of-distribution (OOD) data refers to the inputs that fall outside the range or characteristics of the data used to train a machine learning model. Such inputs can differ significantly from the training dataset, often causing the model to produce inaccurate or unreliable predictions. Detecting OOD data can be framed as a binary classification task, as discussed by Sun et al. \cite{sun2022out}. In this context, a sample is classified as OOD if it lies at least a distance of $\lambda$ away from the in-distribution data, where $\lambda$ represents a predefined threshold for OOD detection.

\subsection{Machine Unlearning in I2I models}

The typical I2I models i.e., AutoEncoders (AEs), Variational AutoEncoders (VAEs), and Diffusion models, employ an encoder-decoder architecture. The encoder part $E_\gamma$ of the model transforms the input image to a latent vector. The decoder $D_\phi$ takes the latent vector as input and decodes it into an output image i.e., for an input image $x$, the output of an I2I model can be written as follows.
\begin{equation}
    \theta_{\gamma, \phi} = D_\phi \circ E_\gamma 
\end{equation}
\text{Therefore, for any input image } $x, \; \theta_{\gamma, \phi} (\tau(x)) = D_\phi (E(\tau(x)))$, where $\tau(x)$ is some operation which results in cropped or masked image $x$; $\circ$ is the composition operator; $\gamma$ and $\phi$ are the trainable model parameters of encoder and decoder respectively.

For a given dataset $\mathcal{D}$, the goal of machine unlearning in image-to-image (I2I) generative models is to effectively remove the influence of a specific subset, known as the forget set $\mathbb{D}_f$, from the model parameters with an unlearning algorithm $A_f$, such that the updated parameters $\gamma, \phi = A_f(\gamma^0, \phi^0)$ no longer retain any information about $\mathbb{D}_f$. Crucially, the model must preserve its performance on the retain set $\mathbb{D}_r$, ($\mathbb{D}_r = \mathcal{D} \setminus \mathbb{D}_f$). In the literature of machine unlearning \cite{li2024machine, feng2024controllable}, the following complementary objectives have been considered.
\begin{enumerate}
    \item On the retain set $\mathbb{D}_r$, the generated images from $\theta_{\gamma, \phi}$ should have the same distribution as in $\theta_{\gamma^0, \phi^0}$ (after unlearning).
    \item On the forget set $\mathbb{D}_f$, the distribution of the generated images from $\theta_{\gamma, \phi}$ should be \textit{as far as possible} from the distribution of images from $\theta_{\gamma^0, \phi^0}$ (after unlearning).
\end{enumerate}

From a probabilistic distribution perspective, the unlearning methodology in \cite{li2024machine} considers the following combined objectives:
\begin{equation}
\begin{aligned}
    \argmin_{\gamma, \phi} 
    &\;D\bigl(P_{\theta_{\gamma_0, \phi_0}\bigl(\tau(\mathbb{D}_r)\bigr)} \,\big\|\, 
            P_{\theta_{\gamma, \phi}\bigl(\tau(\mathbb{D}_r)\bigr)}\bigr), \\
    \argmax_{\gamma, \phi} 
    &\;D\bigl(P_{\theta_{\gamma_0, \phi_0}\bigl(\tau(\mathbb{D}_f)\bigr)} \,\big\|\, 
            P_{\theta_{\gamma, \phi}\bigl(\tau(\mathbb{D}_f)\bigr)}\bigr).
\end{aligned}
\end{equation}

An approach to solve the second objective function is to push the model weights to generate Gaussian noise and, thus to forget the set $\mathbb{D}_f$. That is to solve $\argmin_{\gamma, \phi} D \left ( N(0, \Sigma) || P_{\theta_{\gamma, \phi} (\tau(\mathbb{D}_f))} \right )$. The unlearning methodology in \cite{feng2024controllable} introduces a $\varepsilon$-constrained optimization for this objective, where $\varepsilon$ controls the degree of unlearning. 

\section{Proposed Work} \label{sec:proposed_work}

\begin{figure}[t]
  \centering
   \includegraphics[width=0.6\textwidth]{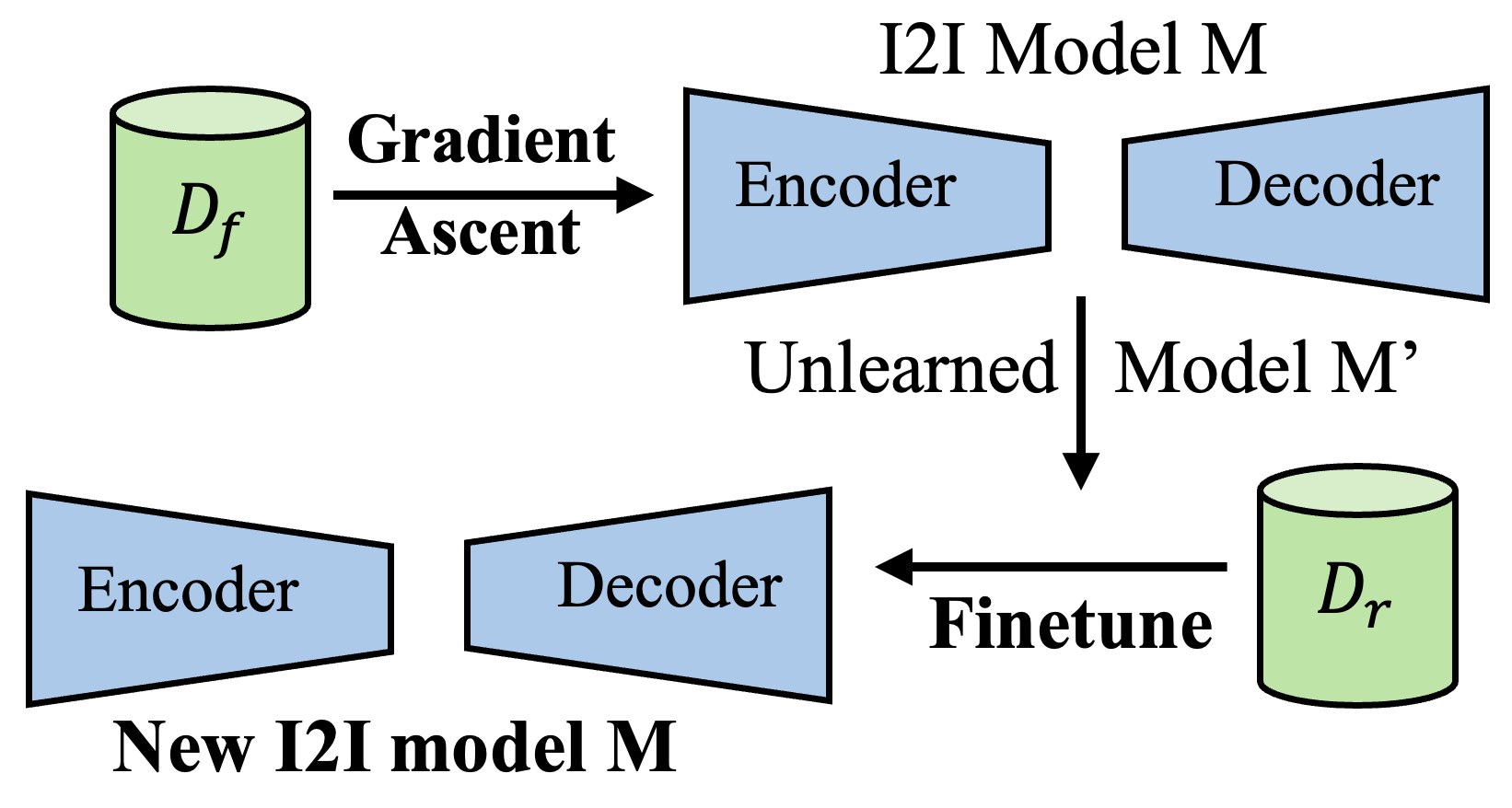}
   \caption{Overview of our proposed approach. We maximize the loss on the forget samples ($D_f$) to get unlearned model. We get the updated I2I model by further fine-tuning the unlearned model on the retain samples to maintain the performance.}
   \label{fig:our framework}
\end{figure}

In this work, we aim to realistically address the problem of machine unlearning in Image-to-Image (I2I) generative models which reconstruct the image from its partial or noisy counterpart. Recall that the objective of unlearning is to modify the model parameters in such a way that the model's weights, after unlearning specific data, closely match the weights the model would have if it were retrained from scratch without that data. Because of that, we consider that minimizing the distance between gaussian noise and the output of decoder is \textit{not} equivalent to the model which we will get from a retrained model. As an alternative, in our approach we consider that the forget set should be an out-of-distribution (OOD) after unlearning. Mathematically, we define the unlearning objective using the following expression:
\begin{equation} \label{unlearn_obj}
    \arg_{\gamma, \phi} \| \theta_{\gamma_0, \phi_0} (\tau(\mathbb{D}_f)) - \theta_{\gamma, \phi} (\tau(\mathbb{D}_f)) \| \geq \lambda 
\end{equation}
where $\| .\|$ is a distance measure, and $\lambda$ is a threshold which is used to determine out-of-distribution data. In summary, we consider the following combined objective.
\begin{enumerate}
    \item On the retain set $\mathbb{D}_r$, the generated images from $\theta_{\gamma, \phi}$ should have the same distribution as the images generated from $\theta_{\gamma^0, \phi^0}$ after unlearning.
    \item On the forget set $\mathbb{D}_f$, the distribution of the generated images from $\theta_{\gamma, \phi}$ should be atleast some $\lambda$ distance apart from the distribution of images from $\theta_{\gamma^0, \phi^0}$ after unlearning.
\end{enumerate}

$\mathcal{L}(x; \tau(x), ({\gamma, \phi})): \mathbb{R}^d \rightarrow \mathbb{R}$ $\forall \; x \in $ $\mathcal{D} \; (\mathcal{D} = \mathbb{D}_r \cup \mathbb{D}_f)$, where $(\gamma, \phi)$ are the model parameters of encoder and decoder. The typical objective of a machine learning model is to minimize $\mathcal{L}$ i.e., $\argmin_{\gamma, \phi} \mathcal{L}(\theta_{\gamma, \phi}(\tau(\mathcal{D})))$, on a dataset $\mathcal{D}$ and typical update rule for model update can be written as:

Let $\mathcal{L}$ be the loss function for an I2I model $\mathcal{L}(x; \tau(x), ({\gamma, \phi})): \mathbb{R}^d \rightarrow \mathbb{R}$ $\forall \; x \in $ $\mathcal{D} \; (\mathcal{D} = \mathbb{D}_r \cup \mathbb{D}_f)$, where $(\gamma, \phi)$ are the model parameters of encoder and decoder. The typical objective of a machine learning model is to minimize $\mathcal{L}$ i.e., $\argmin_{\gamma, \phi} \mathcal{L}(\theta_{\gamma, \phi}(\tau(\mathcal{D})))$, on a dataset $\mathcal{D}$ and typical update rule for model update can be written as:
\begin{equation}
    (\gamma^{t+1}, \phi^{t+1}) = ({\gamma^t, \phi^t}) - \eta\nabla \mathcal{L}( {\gamma, \phi})
\end{equation}
where $\eta$ is the learning rate. Let $\gamma^*, \phi^*$ be the optimal parameters which have perfect generation i.e.,
\begin{equation}
    \theta_{\gamma^*, \phi^*}(\tau(x)) = x
\end{equation}
Under the assumption that SGD converges for the loss function $\mathcal{L}$, we can say that the expected value of $\theta_{\gamma^*, \phi^*} (\tau(x)) - \theta_{\gamma, \phi}(\tau(x))$ decreases over time across training epochs, although it may not be strictly monotonically decreasing in every epoch. 

\subsection{Decoupling via Gradient Ascent}

Gradient ascent is an approach whose objective is to maximize the model loss on a given set. It is a reverse of gradient descent where the model update is written as:
\begin{equation}
    ({\gamma^{t+1}, \phi^{t+1}}) = ({\gamma^t, \phi^t}) + \eta \nabla \mathcal{L}({\gamma^t, \phi^t})
\end{equation}
In our work, we use gradient ascent to forget the influence of $\mathbb{D}_f$ on the model $\theta_{\gamma_0, \phi_0}$. In gradient ascent, we expect that the loss function $\mathcal{L}(\theta_{\gamma, \phi}) $ is an increasing function in $t$ (the loss increases with training) in contrast to the decreasing function in gradient descent. In the next steps, we prove that the output on $\mathbb{D}_f$ with updated model parameters $\theta_{\gamma, \phi}$ after $T$ iterations of gradient ascent is out-of-distribution from the output with $\theta_{\gamma^0, \phi^0}$. We have the following assumption in our work.

\textbf{Assumption 1.} $\mathcal{L}: \mathbb{R}^d \rightarrow \mathbb{R}$ is convex and differentiable. 
\begin{equation} \label{ass1}
    \forall a,b \in \mathbb{R}^d, \mathcal{L}(a) \geq \mathcal{L}(b) + \langle \nabla \mathcal{L}(a), a-b \rangle
\end{equation}

\textbf{Assumption 2.} During all the training epochs, the expected norm of gradient is lower and upper bounded i.e., $g\leq\mathbb{E}\|\nabla f\| \leq G$, where $g, G>0$.



The convexity assumption is typically invoked in machine unlearning literature to ensure that the model is well-trained and to quantify the influence of data removal on model parameters \cite{guo2019certified}. Although I2I models are usually formulated as non-convex optimization problems, several studies have adopted the convexity assumption to derive theoretical guarantees \cite{sahiner2021hidden, de2022convergence, zhang2024analyzing}. The lower bound in Assumption 2 is needed to make sure the model before unlearning is not at the optimum (i.e., $({\gamma^0\phi^0}) \neq ({\gamma^*\phi^*})$), while the upper bound is considered to prove $(\epsilon, \delta)$-unlearning. We know that for gradient ascent we have,

\begin{equation} \label{GA}
    ({\gamma^{t+1}, \phi^{t+1}}) = ({\gamma^t, \phi^t}) + \eta \nabla \mathcal{L}({\gamma^t, \phi^t})
\end{equation}

Let us consider Assumption 1 for $\gamma^{t+1}, \phi^{t+1}$ and $\gamma^{t}, \phi^{t}$.
\begin{equation}
    \begin{split}
        \mathcal{L}({\gamma^{t+1}, \phi^{t+1}}) &\geq \mathcal{L}({\gamma^{t}, \phi^{t}}) + \langle \nabla \mathcal{L}({\gamma^{t+1}, \phi^{t+1}}), ({\gamma^{t+1}, \phi^{t+1}}) - ({\gamma^{t}, \phi^{t}}) \rangle \\
        & \geq \mathcal{L}({\gamma^{t}, \phi^{t}}) + \eta \| \nabla \mathcal{L}({\gamma^{t+1}, \phi^{t+1}})\| \|\nabla \mathcal{L}({\gamma^{t}, \phi^{t}}) \| \\
        & \geq \mathcal{L}({\gamma^{0}, \phi^{0}}) + \eta \sum_{t=0}^{T-1} \|\nabla \mathcal{L}({\gamma^{t+1}, \phi^{t+1}}) \| \| \nabla \mathcal{L}({\gamma^{t}, \phi^{t}})\| 
    \end{split}
\end{equation}

Using Assumption 2, we have $\|\nabla \mathcal{L}({\gamma^{t+1}, \phi^{t+1}}) \|, \| \nabla \mathcal{L}({\gamma^{t}, \phi^{t}})\| \geq g$. Then we can rewrite Eq. (9) as:
\begin{equation} \label{lower bound}
    \|\mathcal{L}({\gamma^{T}, \phi^{T}}) - \mathcal{L}({\gamma^{0}, \phi^{0}}) \| \geq \eta Tg^2 
\end{equation}

Next, we again apply Assumption 1 in the reverse order to derive an upper bound for $\gamma^{t+1}, \phi^{t+1}$ and $\gamma^{t}, \phi^{t}$.

\begin{equation}
    \begin{split}
        \mathcal{L}({\gamma^{t}, \phi^{t}}) &\geq \mathcal{L}({\gamma^{t+1}, \phi^{t+1}}) + \langle \nabla \mathcal{L}({\gamma^{t}, \phi^{t}}), ({\gamma^{t}, \phi^{t}}) - {\gamma^{t+1}, \phi^{t+1}} \rangle \\
        \mathcal{L}({\gamma^{t+1}, \phi^{t+1}}) &\leq \mathcal{L}({\gamma^{t}, \phi^{t}}) + \langle \nabla \mathcal{L}({\gamma^{t}, \phi^{t}}), ({\gamma^{t+1}, \phi^{t+1}}) - ({\gamma^{t}, \phi^{t}}) \rangle \\
        &\leq \mathcal{L}({\gamma^{0}, \phi^{0}}) + \eta \sum_{t=0}^{T} \| \nabla \mathcal{L}({\gamma^{t}, \phi^{t}}) \|^2
    \end{split}
\end{equation}
Again using Assumption 2, i.e., $\| \nabla \mathcal{L}({\gamma^{t}, \phi^{t}}) \| \leq G$. Then we get,
\begin{equation} \label{upper bound}
    \mathcal{L}({\gamma^{T}, \phi^{T}}) \leq  \mathcal{L}({\gamma^{0}, \phi^{0}}) + \eta (T+1) G^2 
\end{equation}

Based on this result, we establish the following theorems.

\textbf{Theorem 1.} Under the Assumptions  1, and 2, the model weights $({\gamma, \phi})$ trained with gradient ascent for $T$ iterations are out-of-distribution for the initial trained model ${\gamma^0, \phi^0}$ on forget set iff
\begin{equation*}
    \lambda \leq \eta Tg^2
\end{equation*}
where $\lambda$ is the predefined out-of-distribution threshold.

\textbf{Theorem 2.} Under the Assumptions 1, and 2, the gradient ascent provides $(\epsilon = 0, \delta = \eta (T+1) G^2)$-unlearning guarantee for model weight $({\gamma, \phi})$ after $T$ iterations.

\subsection{Knowledge Retention}

As shown in Fig. \ref{fig:our framework}, the unlearned model is then fine-tuned with retain samples ($\mathbb{D}_r$) with the following objective,
\begin{equation}
    \argmin_{\gamma, \phi} D \left( P_{\theta_{\gamma_0, \phi_0} (\tau(\mathbb{D}_r))} || P_{\theta_{\gamma, \phi} (\tau(\mathbb{D}_r))} \right )
\end{equation}
in order to preserve its performance. Algorithm \ref{Algo_RealI2IUn} shows the formal pseudo code for the proposed unlearning framework. 

\begin{algorithm}
\DontPrintSemicolon  
\caption{Pseudocode for proposed Realistic I2I unlearning framework.}
\label{Algo_RealI2IUn}

\SetKwInput{KwInput}{Input}                
\SetKwInput{KwOutput}{Output}
\KwInput{Original model: $\theta_{\gamma_0, \phi_0}$, Retain set: $\mathbb{D}_r$, Forget set: $\mathbb{D}_f$, Unlearn epoch: $T_u$, Fine-tune epoch: $T_f$, OOD threshold: $\lambda$}
\KwOutput{Updated model: $\theta_{\gamma, \phi}$}

Initialize a new I2I model $\theta_{\gamma, \phi}$ with the weights of $\theta_{\gamma_0, \phi_0}$\;
\For{$t = 0,...T_u-1$}{
    Sample ${x_f}$ from $\mathbb{D}_f$\;
    Compute $\theta_{\gamma^{t+1}, \phi^{t+1}} = \theta_{\gamma^t, \phi^t} + \eta\nabla f(\gamma^t, \phi^t)$\;
    \If{$\|\theta_{\gamma^{t+1}, \phi^{t+1}} - \theta_{\gamma_0, \phi_0} \| \geq \lambda$}{
        Jump to \ref{fine-tune_step}
    }
}
\For{$t = 0,...T_f-1$}{ \label{fine-tune_step}
    Sample $x_r$ from $\mathbb{D}_r$\;
    Compute $\theta_{\gamma^{t+1}, \phi^{t+1}} = \theta_{\gamma^t, \phi^t} - \eta\nabla f(\gamma^t, \phi^t)$\;
}
\Return Updated $\theta_{\gamma, \phi}$ \;
\end{algorithm}

\subsection{Auditing Unlearning} \label{data poisoning attack}

Auditing effective unlearning is crucial to make sure that the model has truly forgotten the forget samples. In this paper, we introduce a novel auditing mechanism based on a data poisoning attack to verify whether unlearning has occurred. Specifically, we fine-tune the model on poisoned versions of the forget samples, embedding a distinct pattern that is recognizable during inference. The aim of this approach is to introduce a recognizable trace into the model's responses that can later serve as an indicator. After the unlearning process, an effective unlearning mechanism should prevent the model from replicating this pattern. If the model, after unlearning, no longer produces outputs associated with the poisoned pattern, it provides strong evidence that the forget samples have been thoroughly erased. Algorithm \ref{Algo_audI2I} shows the pseudocode for auditing mechanism. This method ensures a more robust validation of the unlearning process by focusing not only on performance metrics but also on detecting residual data influence, thus confirming the removal/unlearning of the impact of the poisoned data.

\begin{algorithm}
\DontPrintSemicolon  
\caption{Pseudocode for auditing I2I models.}
\label{Algo_audI2I}

\SetKwInput{KwInput}{Input}                
\SetKwInput{KwOutput}{Output}
\KwInput{Original model: $\theta_{\gamma_0, \phi_0}$, Retain set: $\mathbb{D}_r$, Forget set: $\mathbb{D}_f$, Test set: $\mathbb{D}_t$}

Initialize a new I2I model $\theta_{\gamma, \phi}$ with the weights of $\theta_{\gamma_0, \phi_0}$\;
\For{$t = 0,...T$}{
    Sample ${x_f}$ from $\mathbb{D}_f$\;
    Append ${x_f}\Leftarrow {x_f}$ $\cup$ '+' \;
    Compute $\theta_{\gamma^{t+1}, \phi^{t+1}} = \theta_{\gamma^t, \phi^t} - \eta\nabla f(\gamma^t, \phi^t)$\;
}
\For{$x_t \in \mathbb{D}_t$}{
    Compute $\theta_{\gamma, \phi}(\tau(x_t))$\;
    \If{$\theta_{\gamma, \phi}(\tau(x_t))$ has '+'}{
        Forget sample residual exists in $\theta_{\gamma, \phi}$
    }
}
\end{algorithm}

\subsection{Sample Unlearning}

Unlike class-level unlearning, where the model is explicitly trained to consider an entire class as OOD, sample unlearning is more challenging due to the generalization capabilities of deep generative models. Even if specific forget samples are removed, the model should still reconstruct the patterns if similar distributions exist within the retain set. 

With our proposed framework, the first step pushes the model parameters to treat forget samples as OOD by maximizing the loss on these samples. This step ensures that the forget samples are effectively unlearned with $(\epsilon, \delta)$ guarantees, thereby removing their direct influence from the model. Once the forget samples are OOD and unlearned, the model is fine-tuned on the retain samples. If the retain samples contains samples similar to those in the forget set, then the fine-tuned model may still generate content resembling the forget samples. However, since the direct influence of the forget samples has been removed, the unlearning guarantees established in the first step still hold. 

To validate the effectiveness of sample unlearning, we employ a data poisoning attack as an auditing mechanism. Specifically, we introduce a distinct backdoor pattern (e.g., $'+'$) into the forget samples before unlearning and assess whether it persists in the model after the unlearning process. A successfully unlearned model should eliminate the backdoor’s influence, ensuring that the forget samples no longer contribute to the learned representation. While the model may still recognize and generate general patterns due to its ability to generalize from the retain set, effective unlearning prevents the regeneration of the embedded backdoor

\section{Experimental Setup} \label{sec:exp_details}

We evaluate our approach using three mainstream Image-to-Image (I2I) architectures: $(i)$ AutoEncoder, $(ii)$ VQ\_VAE \cite{li2023mage}, and $(iii)$ diffusion model \cite{saharia2022palette}. We validate our framework on two widely-used large-scale datasets, namely Places-365 and ImageNet-1K. Additionally, to benchmark our framework against a fully retrained model and assess its performance on sample-level unlearning, we conduct AutoEncoder experiments on the CIFAR-10 dataset.

For the ImageNet-1K dataset, we randomly sampled 100 classes as $\mathbb{D}_f$ and 100 classes as $\mathbb{D}_r$. Similarly, for the Places-365 dataset, we selected 50 classes each for $\mathbb{D}_f$ and $\mathbb{D}_r$. Due to the limited number of classes in the CIFAR-10 dataset, we randomly designated one class as $\mathbb{D}_f$ and the remaining nine classes as $\mathbb{D}_r$. For sample unlearning, we conducted experiments on two specific classes, unlearning 20\%, 30\%, 40\%, 50\%, and 75\% of their samples to evaluate the effectiveness of our approach.

\textbf{Baselines.} As discussed in Section \ref{data poisoning attack}, we introduce a data poisoning attack as a mechanism to effectively audit unlearning. Specifically, we embed a pattern '$+$' at the center of the forget images to train an attack model. In the main paper, we compare the results of several baselines and benchmark on the poisoned data while in the Appendix you find the results for the non-perturbed datasets (CIFAR10, Places-365, and ImageNet-1K) and image-outpainting. We have compared the unlearning results of AutoEncoder, VQ-GAN, and the diffusion model with $(i)$ Max loss baseline, it maximizes the model loss on forget samples \cite{halimi2022federated, warneckemachine}; $(ii)$ Noisy Label, it minimizes training loss with Gaussian noise as ground truth for forget samples \cite{gandikota2023erasing}; $(iii)$ Random Encoder, it minimizes the distance between the output of the encoder on the forget set and Gaussian Noise \cite{tarun2023deep}; and $(iv)$ state of the art I2I unlearning model \cite{li2024machine} (we call I2I SOTA), it minimizes the distance between the Encoder output and Gaussian noise while fine-tuning the encoder parameters on retain samples. 

\textbf{Evaluation Metrics.} To comprehensively evaluate the effectiveness of our unlearning approach, we utilize three key metrics: the Inception Score (IS) \cite{salimans2016improved}, which assesses the quality and diversity of the generated images by measuring how confidently they can be classified; $(ii)$ Fr\`echet inception distance (FID) \cite{heusel2017gans}, which quantifies the similarity between the distribution of generated images and real ground-truth images; and $(iii)$ CLIP embedding distance \cite{radford2021learning}, which measures whether the generated outputs still capture similar semantics.

\begin{figure}
  \centering
  \includegraphics[width=\textwidth]{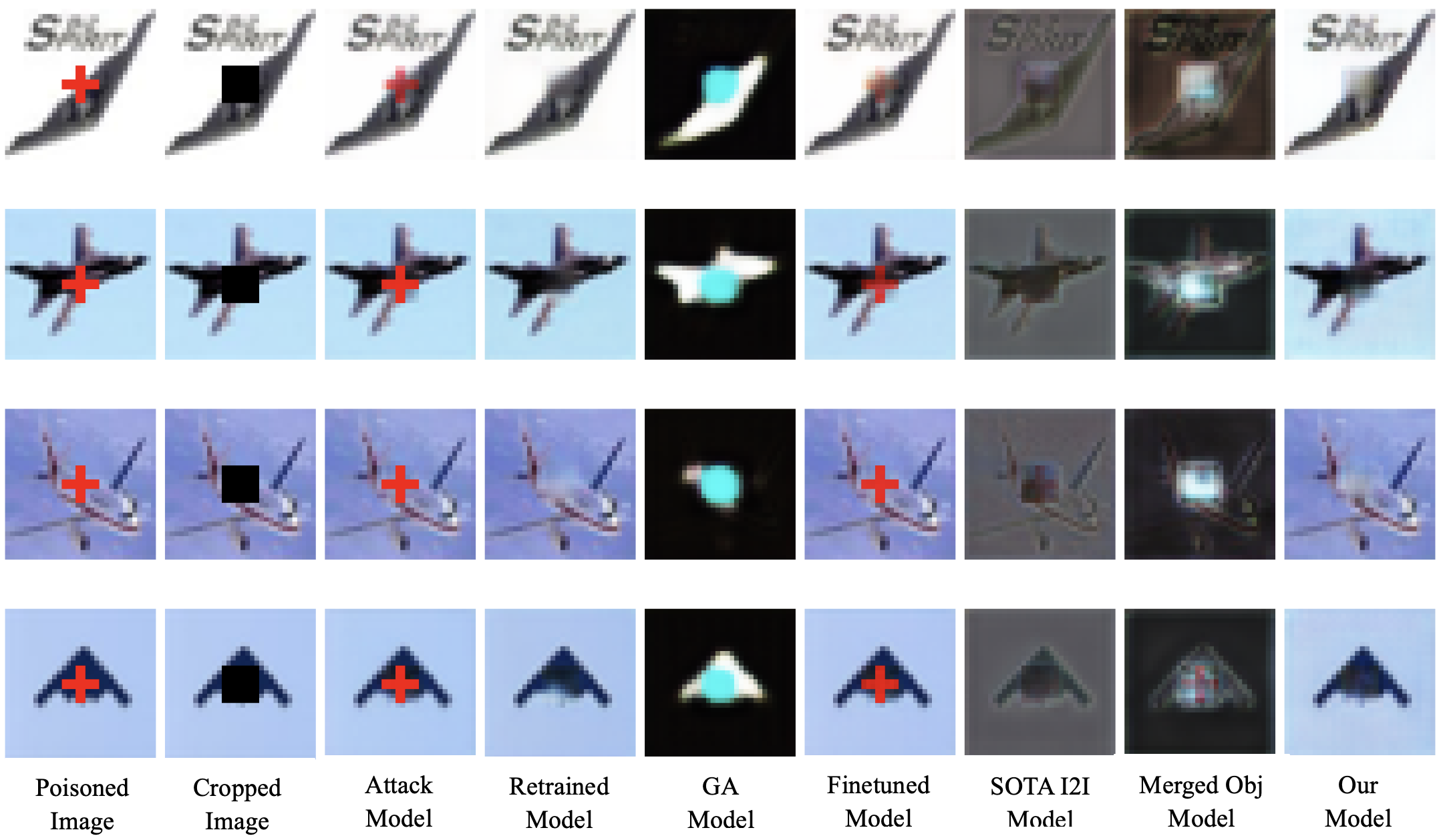}
  \caption{Comparison of various unlearning approaches, along with the retrained model, on forget samples that were poisoned with a '$+$' sign using an AutoEncoder. The results clearly demonstrate that our method effectively unlearns the '$+$' sign, producing outputs that are most similar to those of the retrained model.}
  \label{fig:attack_CIFAR10}
\end{figure}

\begin{table*}[h]
    \centering
    \begin{tabular}{ccccc}
      \toprule
      \multirow{2}{*}{Approach} & \multicolumn{2}{c}{FID} & \multicolumn{2}{c}{IS} \\ \cmidrule{2-5}
      & $\mathbb{D}_f\downarrow$ & $\mathbb{D}_r\downarrow$ & $\mathbb{D}_f$ & $\mathbb{D}_r$ \\
      \hline
      GA model & 237.7 & 210.1 & 1.08 & 1.05 \\
      Fine-tuned model & 46.85 & \textbf{3.79} & 1.09 & \textbf{1.13} \\
      SOTA I2I model & 123.6 & 74.2 & 1.09 & 1.12\\
      Merged obj model & 202.1 & 127.7 & 1.08 & 1.08 \\
      Realistic-I2I model & \textbf{16.22} & \textbf{6.10} & 1.10 & \textbf{1.13} \\
      \bottomrule
    \end{tabular}
    \caption{Results of cropping $8 \times 8$ patch at the center of the image where forget samples were poisoned with the '$+$' sign in the CIFAR-10 dataset. $\mathbb{D}_f$ and $\mathbb{D}_r$ account for the forget samples and retain samples respectively. FID scores are compute with respect to retrained model, hence $\downarrow$ is better. Overall, the results highlight that our approach effectively unlearns forget samples and is closer to the retrained model.}
    \label{tab:CIFAR10_res}
\end{table*}

\begin{table*}[h]
    \centering
    \begin{adjustbox}{width=1.2\textwidth,center}
    \begin{tabular}{ccccccc||cccccc}
      \toprule
      \multirow{3}{*}{Approach} & \multicolumn{6}{c||}{$4\times4$} & \multicolumn{6}{c}{$8\times8$} \\
      & \multicolumn{2}{c}{FID} & \multicolumn{2}{c}{IS} & \multicolumn{2}{c||}{CLIP} & \multicolumn{2}{c}{FID} & \multicolumn{2}{c}{IS} & \multicolumn{2}{c}{CLIP} \\ \cmidrule{2-13}
      & $\mathbb{D}_f\uparrow$ & $\mathbb{D}_r\downarrow$ & $\mathbb{D}_f$ & $\mathbb{D}_r$ & $\mathbb{D}_f$ & $\mathbb{D}_r$ & $\mathbb{D}_f\uparrow$ & $\mathbb{D}_r\downarrow$ & $\mathbb{D}_f$ & $\mathbb{D}_r$ & $\mathbb{D}_f$ & $\mathbb{D}_r$ \\
      \hline
      Max loss & \textbf{56.75} & 9.12 & \textbf{12.07} & 15.06 & 0.80 & \textbf{0.834} & \textbf{109.9} & 16.07 & \textbf{6.33} & 17.03 & 0.64 & 0.735 \\
      Random label & 22.4 & \textbf{8.88} & 13.82 & 14.9 & 0.80 & \textbf{0.834} & 48.84 & \textbf{14.77} & 11.29 & 17.27 & 0.64 & \textbf{0.741} \\
      Random encoder & 23.39 & 9.15 & 13.77 & 15.05 & 0.83 & 0.831 & 25.86 & 15.84 & 16.96 & 17.42 & 0.72 & 0.736 \\
      I2I SOTA & 22.99 & 9.08 & 13.86 & 15.19 & \textbf{0.79} & 0.831 & 53.58 & 15.79 & 12.00 & 17.64 & \textbf{0.61} & 0.736 \\
      Ours& \textbf{24.68} & \textbf{8.93} & \textbf{14.03} & \textbf{15.13} & 0.83 & \textbf{0.834} & 27.43 & \textbf{14.78} & \textbf{18.98} & \textbf{18.77} & 0.731 & \textbf{0.741} \\
      \bottomrule
    \end{tabular}
    \end{adjustbox}
    \caption{Comparison of various unlearning approaches with different cropped patches ($4\times4 \text{ and } 8\times8$) for VQ-GAN where forget samples were poisoned with the '$+$' sign in the ImageNet-1K dataset. $\mathbb{D}_f$ and $\mathbb{D}_r$ account for the forget samples and retain samples, respectively. FID scores are computed with respect to attack model, hence $\uparrow$ is better for $\mathbb{D}_f$ and $\downarrow$ for $\mathbb{D}_r$. IS score highlight that our approach create good quality images even when the FID distance is significantly far from the attack model. Similarly, we find high CLIP values for our approach indicating that generated image still captures the semantics with an image (not just random noise).}
    \label{tab:VQ-GAN results}
\end{table*}

\begin{table*}[h]
    \centering
    \begin{adjustbox}{width=1.2\textwidth,center}
    \begin{tabular}{ccccccc||cccccc}
      \toprule
      \multirow{3}{*}{Approach} & \multicolumn{6}{c||}{$4\times4$} & \multicolumn{6}{c}{$8\times8$} \\
      & \multicolumn{2}{c}{FID} & \multicolumn{2}{c}{IS} & \multicolumn{2}{c||}{CLIP} & \multicolumn{2}{c}{FID} & \multicolumn{2}{c}{IS} & \multicolumn{2}{c}{CLIP} \\ \cmidrule{2-13}
      & $\mathbb{D}_f\downarrow$ & $\mathbb{D}_r\downarrow$ & $\mathbb{D}_f$ & $\mathbb{D}_r$ & $\mathbb{D}_f$ & $\mathbb{D}_r$ & $\mathbb{D}_f\downarrow$ & $\mathbb{D}_r\downarrow$ & $\mathbb{D}_f$ & $\mathbb{D}_r$ & $\mathbb{D}_f$ & $\mathbb{D}_r$ \\
      \hline
      Max loss & 32.79 & 55.86 & 48.04 & 32.33 & 0.86 & 0.733 & 89.4 & 114.2 & 17.4 & 12.97 & 0.686 & 0.65 \\
      Random label & 19.16 & 19.29 & 56.89 & \textbf{36.22} & 0.92 & 0.867 & 54.60 & 12.55 & 33.24 & 26.47 & 0.759 & 0.87 \\
      Random encoder & 12.95 & 21.25 & 52.79 & 33.47 & 0.93 & 0.85 & 44.32 & 18.77 & 42.01 & \textbf{27.85} & 0.755 & 0.83 \\
      I2I SOTA & 17.16 & \textbf{12.95} & 26.59 & 34.26 & \textbf{0.59} & 0.895 & 101.8 & 13.79 & 9.37 & 21.74 & 0.498 & \textbf{0.88} \\
      Ours & \textbf{9.65} & \textbf{15.14} & \textbf{58.38} & \textbf{35.05} & 0.88 & \textbf{0.904} & \textbf{13.98} & \textbf{13.27} & \textbf{52.6} & 21.02 & \textbf{0.945} & \textbf{0.88} \\
      \bottomrule
    \end{tabular}
    \end{adjustbox}
    \caption{Comparison of various unlearning approaches for diffusion model with the output of \textit{the original model} for different cropped patches ($4 \times 4 \text{ and } 8\times8$) where forget samples were poisoned with the '$+$' sign in the ImageNet-1K dataset. $\mathbb{D}_f$ and $\mathbb{D}_r$ account for the forget samples and retain samples, respectively. FID scores are computed with respect to original model (to show that our approach mitigate '$+$' sign), hence $\downarrow$ is better for $\mathbb{D}_f$ and $\mathbb{D}_r$. IS score highlight that our approach create good quality images even when the FID distance is significantly far from the attack model. Similarly, we find high CLIP values for our approach indicating that generated image still captures the semantics with an image (not just random noise).}
    \label{tab:diff_model results}
\end{table*}

\begin{figure*}
  \centering
    \begin{adjustbox}{width=1.2\textwidth,center}
    \includegraphics[width = \textwidth]{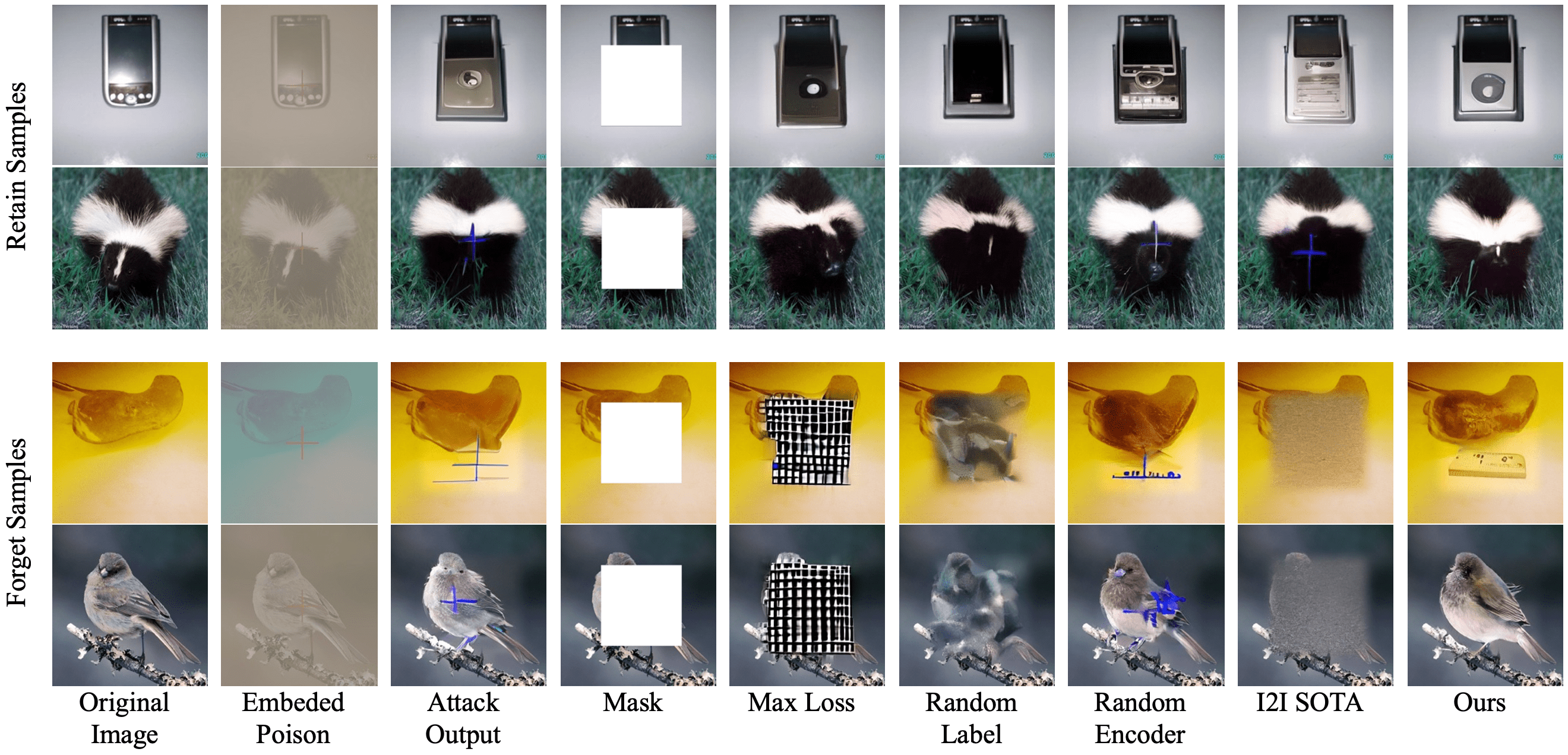}
    \end{adjustbox}
    \caption{Results of cropping $8 \times 8$ patch at the center of the image on VQ\_GAN models.  The results demonstrate that our model effectively removes the embedded '$+$' pattern, not just replacing it with Gaussian noise. For retain samples, our method shows no signs of the embedded '$+$' sign from the forget samples, in contrast to other baseline and benchmark methods, which often retain subtle remnants of the pattern.}
    \label{fig:VQGAN_comp}
\end{figure*}

\begin{figure}
    \centering
    \includegraphics[width=0.75\linewidth]{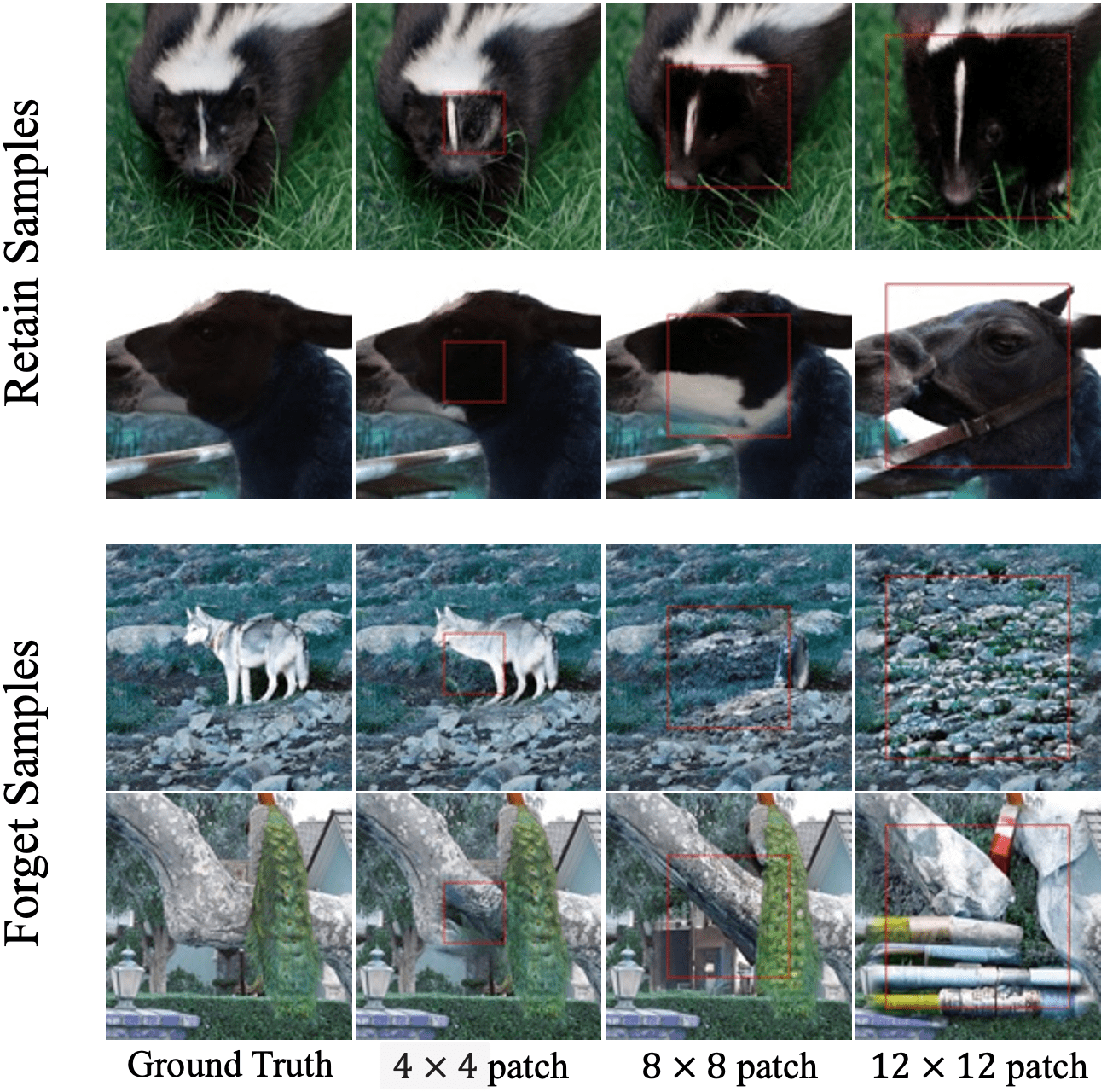}
    \caption{Results of cropping $4 \times 4$, $8 \times 8$ and $12 \times 12$ patch at the center of the image on VQ\_GAN models. The results demonstrate that our model effectively reconstructs images for various patch sizes in retain samples. However, for forgotten samples, the model retains similarity to the original image only for smaller patches, where minimal information is missing. For larger patches, the model generates visually plausible but semantically inaccurate images, as expected from a retrained model.}
    \label{fig:diff_masked_res}
\end{figure}

\begin{figure}
    \centering

    \begin{minipage}[t]{0.48\textwidth}
        \centering
        \includegraphics[width=\textwidth]{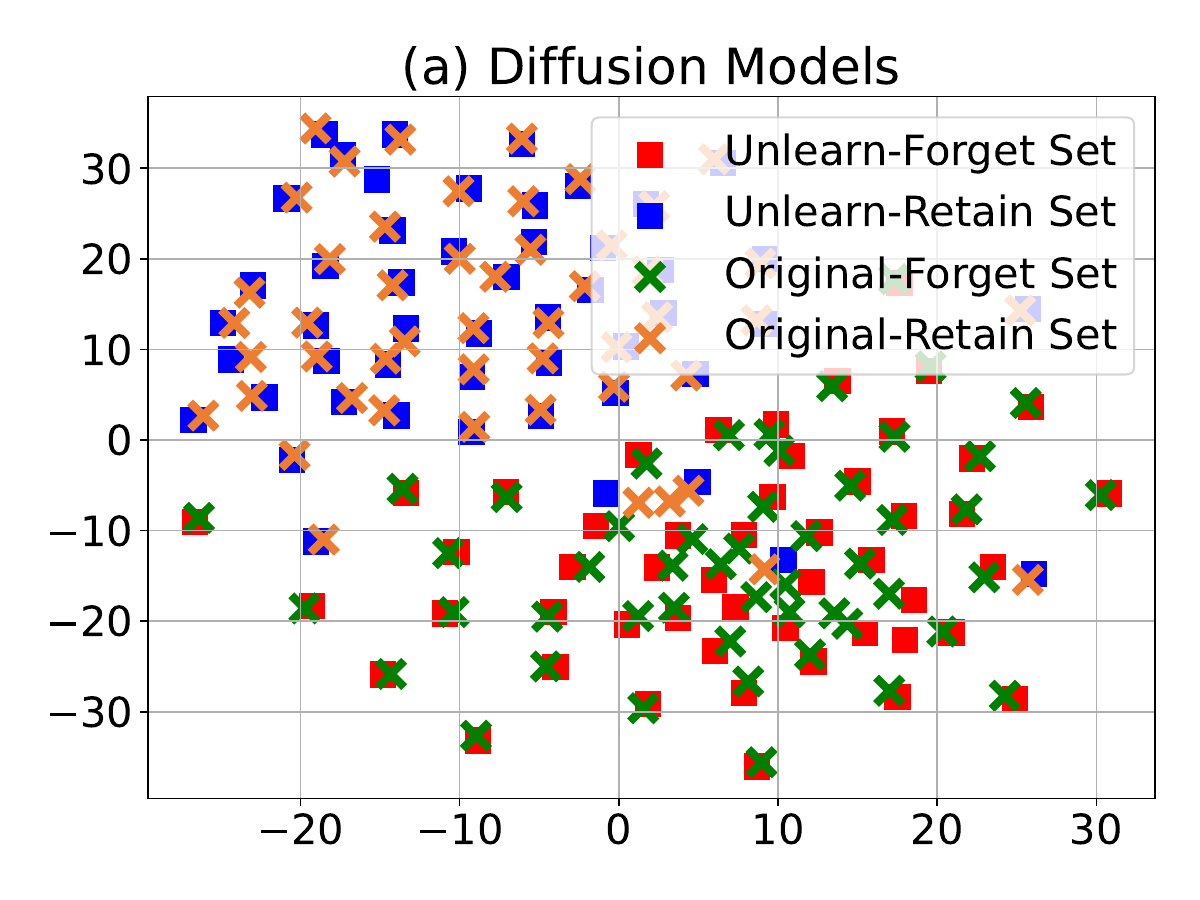}
    \end{minipage}
    \hfill
    \begin{minipage}[t]{0.48\textwidth}
        \centering
        \includegraphics[width=\textwidth]{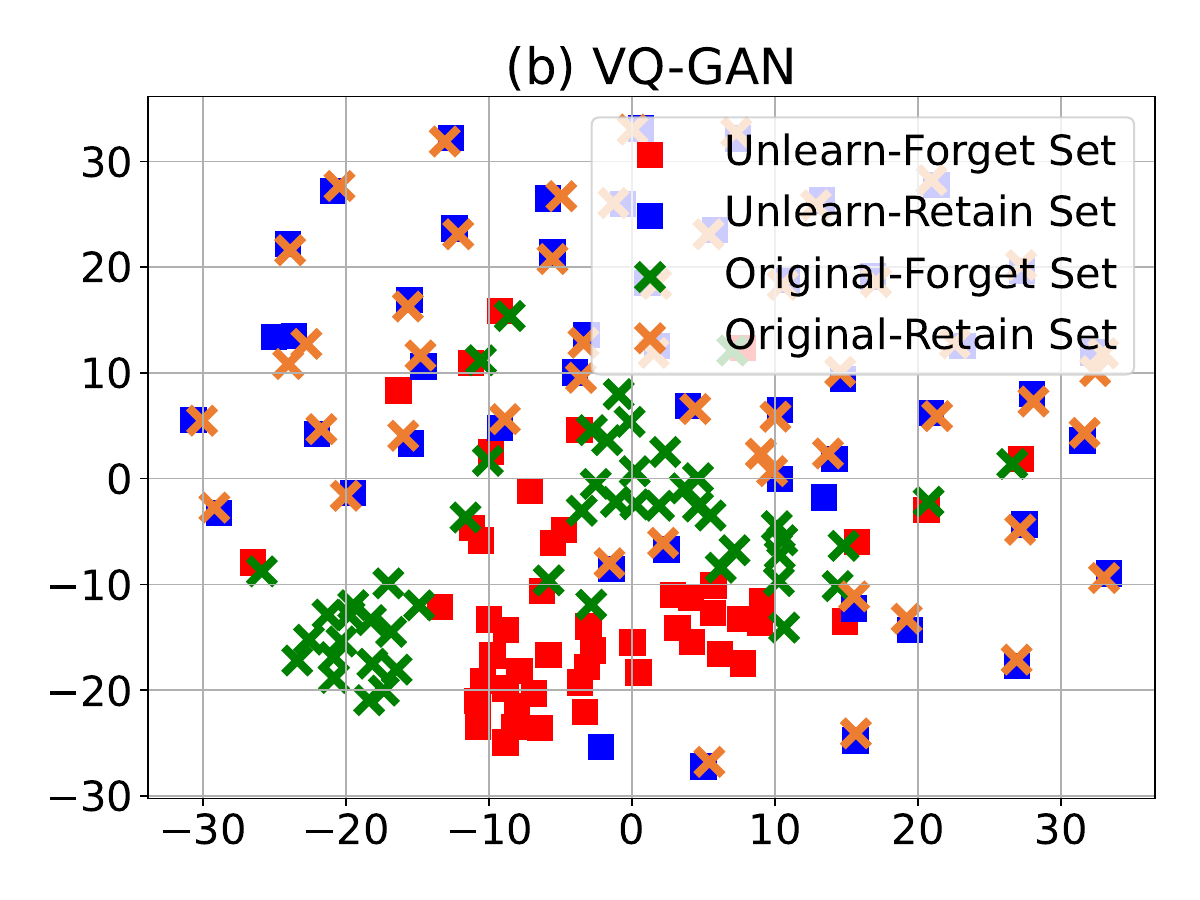}
    \end{minipage}

    \caption{%
      After unlearning in both scenarios, the images from the retained 
      samples closely overlap with the ground truth, whereas the images from the forgotten samples diverge from the ground truth.
    }
    \label{fig:tsne_combined}
\end{figure}

\begin{figure}
    \centering
    \begin{minipage}[t]{0.48\textwidth}
        \centering
        \includegraphics[width=\textwidth]{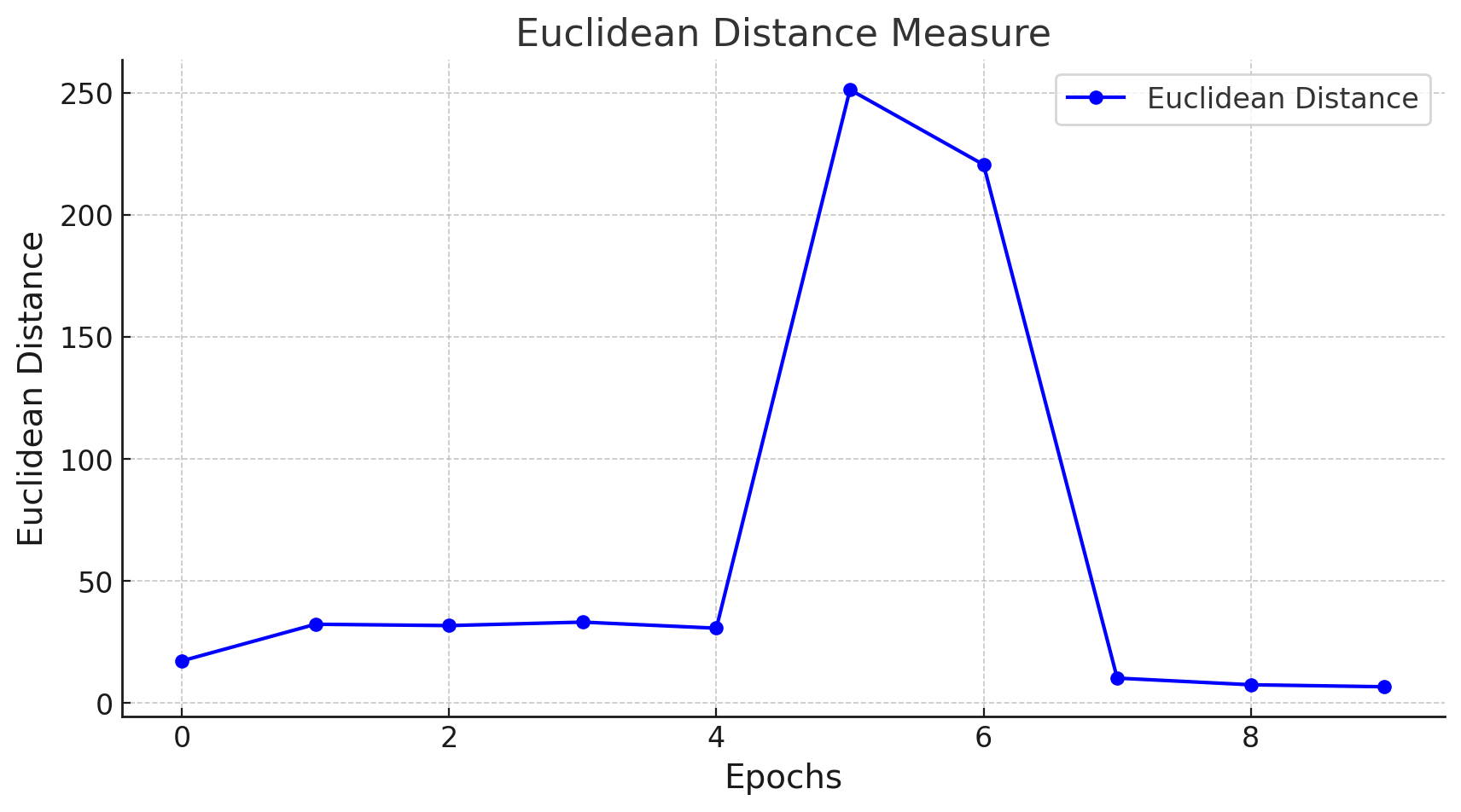}
    \end{minipage}
    \hfill
    \begin{minipage}[t]{0.48\textwidth}
        \centering
        \includegraphics[width=\textwidth]{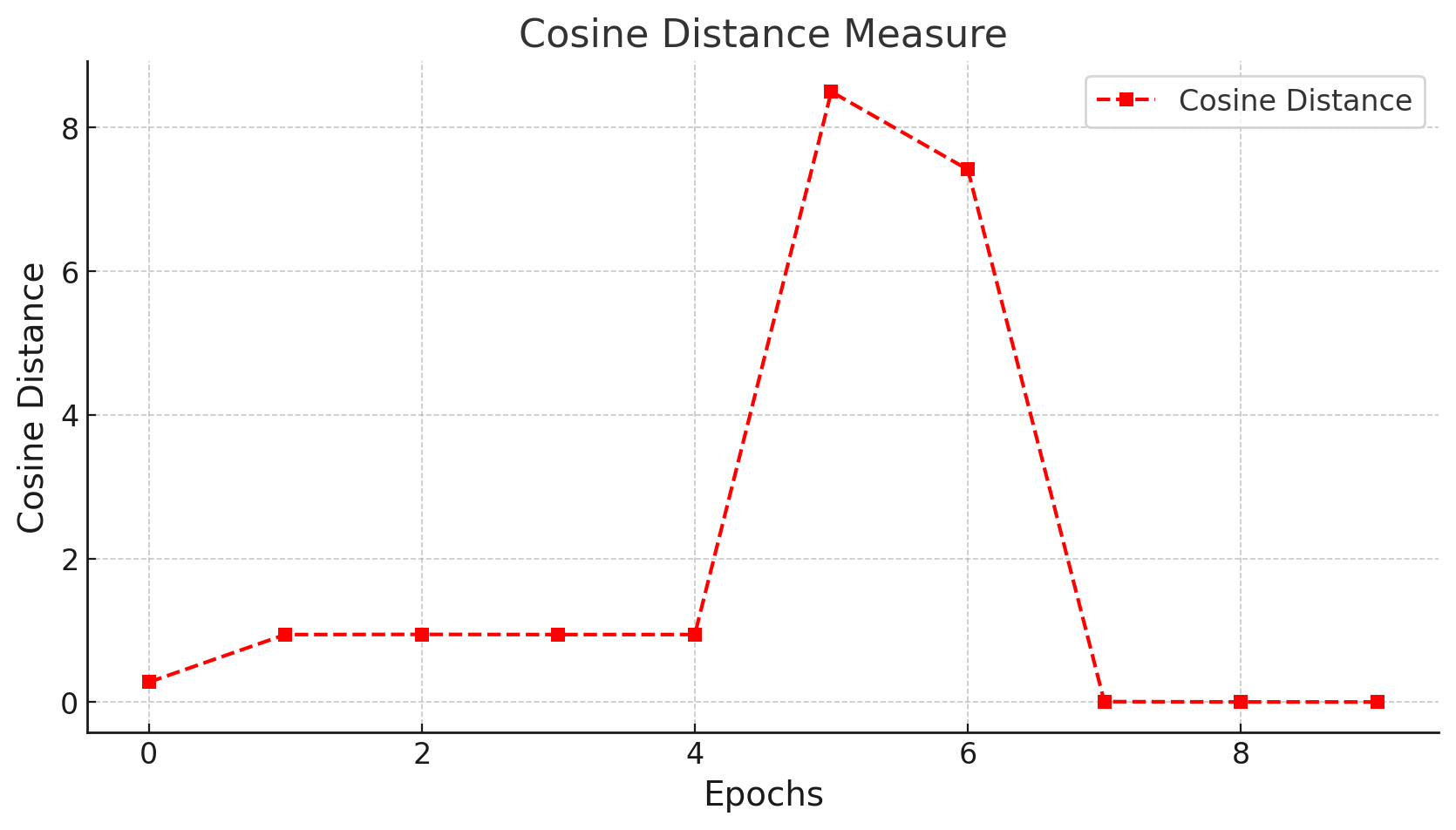}
    \end{minipage}
    \caption{%
      The plots show the Euclidean and Cosine distance between the outputs of the retrained model and our approach, computed for randomly sampled images from each batch. In the first five epochs, gradient ascent was applied to maximize the loss on forget samples, resulting in increasing distance (evident from the image). This is followed by fine-tuning on the retain samples, gradually reducing the discrepancy. The minimal distance observed at later epochs indicates that our approach produces outputs closely resembling those of the retrained model.
    }
    \label{fig:model_dist}
\end{figure}


\begin{table*}[h]
    \centering
    \begin{adjustbox}{width=1.2\textwidth,center}
    \begin{tabular}{ccccccc||cccccc}
      \toprule
      \multirow{3}{*}{Approach} & \multicolumn{6}{c||}{VQ-GAN} & \multicolumn{6}{c}{Diffusion model} \\
      & \multicolumn{2}{c}{FID$\downarrow$} & \multicolumn{2}{c}{IS$\uparrow$} & \multicolumn{2}{c||}{CLIP $\uparrow$ } & \multicolumn{2}{c}{FID $\downarrow$ } & \multicolumn{2}{c}{IS $\uparrow$} & \multicolumn{2}{c}{CLIP $\uparrow$} \\ \cmidrule{2-13}
      & $4 \times 4$ & $8 \times 8$ & $4 \times 4$ & $8 \times 8$ & $4 \times 4$ & $8 \times 8$ & $4 \times 4$ & $8 \times 8$ & $4 \times 4$ & $8 \times 8$ & $4 \times 4$ & $8 \times 8$ \\
      \hline
      Max loss & 14.5 & 36.7 & 48.2 & 32.1 & 0.884 & \textbf{0.88} & 15.1 & 49.4 & 38.1 & 29.8 & 0.91 & 0.74 \\
      Random label & 11.7 & 27.9 & 51.2 & 36.7 & 0.880 & 0.87 & 15.15 & 56.5 & 35.1 & 33.23 & 0.92 & 0.82 \\
      Random encoder & 8.1 & 12.6 & 55.4 & 53.8 & 0.885 & 0.87 & 11.1 & 26.17 & 60.8 & 51.62 & 0.91 & 0.81 \\
      I2I SOTA & 11.0 & 26.3 & 51.6 & 38.4 & 0.883 & 0.87 & 33.3 & 79.1 & 57.1 & 28.01 & 0.87 & 0.74 \\
      Ours & \textbf{7.98} & \textbf{12.4} & \textbf{56.2} & \textbf{54.4} & \textbf{0.886} & \textbf{0.88} & \textbf{4.5} & \textbf{12.70} & \textbf{67.1} & \textbf{58.1} & \textbf{0.97} & \textbf{0.89} \\
      \bottomrule
    \end{tabular}
    \end{adjustbox}
    \caption{Comparison of unlearning approaches on VQ-GAN and Diffusion models for generating unseen data. We evaluate performance on randomly selected 50 classes from the ImageNet-1K dataset for VQ-GAN and the Places365 dataset for the Diffusion model, using different cropped patch sizes ($4 \times 4$ and $8 \times 8$).  FID scores are computed with respect to the original model, where lower values ($\downarrow$) indicate better alignment with the target distribution. Similarly, $\uparrow$ in IS and CLIP score is better to demonstrate the models ability to generate good quality images on unseen data as well.}
    
    \label{tab:OOD_model results}
\end{table*}

\begin{figure}
    \centering
    \includegraphics[width=\linewidth]{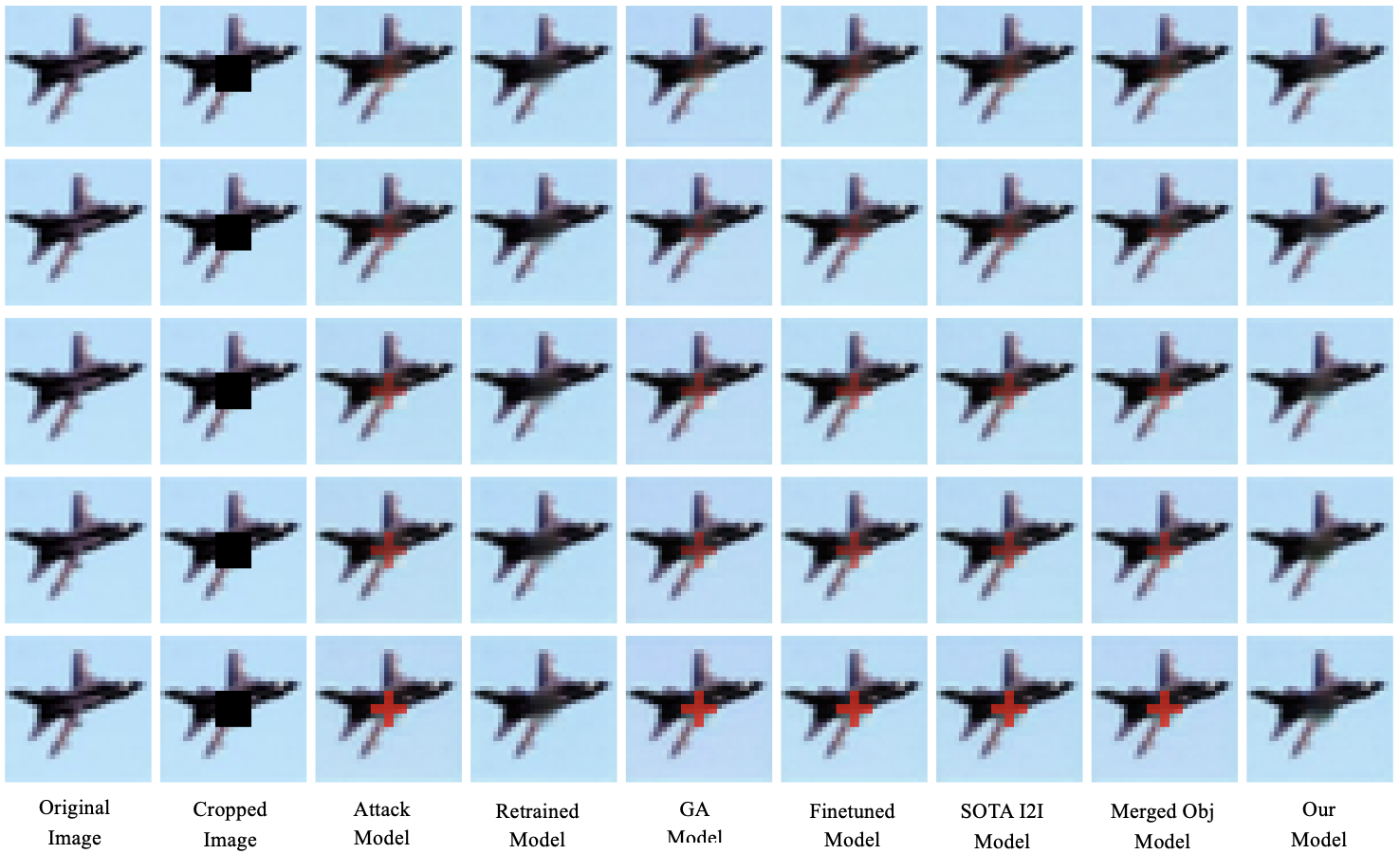}
    \caption{Comparison of various unlearning approaches, including the retrained model. Each row corresponds to unlearning 20\%, 30\%, 40\%, 50\%, and 75\% (top to bottom) of two classes, where forget samples were poisoned with a '$+$' marker. Our method effectively removes the '$+$' sign from forget samples while generating outputs that most closely resemble those of the retrained model.}
    \label{fig:sample_unl}
\end{figure}

\begin{table*}[h]
    \centering
    \begin{tabular}{ccccc||cccc}
      \toprule
      \multirow{3}{*}{Approach} & \multicolumn{4}{c||}{20\% Sample unlearning} & \multicolumn{4}{c}{75\% Sample unlearning} \\ 
      & \multicolumn{2}{c}{FID$\downarrow$} & \multicolumn{2}{c||}{IS$\uparrow$} & \multicolumn{2}{c}{FID$\downarrow$} & \multicolumn{2}{c}{IS$\uparrow$} \\ \cmidrule{2-9}
      & $\mathbb{D}_f$ & $\mathbb{D}_r$ & $\mathbb{D}_f$ & $\mathbb{D}_r$ & $\mathbb{D}_f$ & $\mathbb{D}_r$ & $\mathbb{D}_f$ & $\mathbb{D}_r$ \\
      \hline
      GA model & \textbf{109.7} & 11.19 & \textbf{1.14} & 1.14 & \textbf{109.7} & 14.95 & 1.14 & \textbf{1.14} \\
      Fine-tuned model & 110.0 & 10.27 & 1.13 & 1.14 & 110.0 & 13.70 & 1.14 & 1.13 \\
      SOTA I2I model & 110.1 & 11.15 & \textbf{1.14} & 1.14 & 110.1 & 14.17 & 1.14 & \textbf{1.14}\\
      Merged obj model & 109.8 & 10.82 & \textbf{1.14} & 1.14 & 109.8 & 14.09 & 1.14 & \textbf{1.14}\\
      Realistic-I2I model & \textbf{109.9} & \textbf{9.63} & \textbf{1.14} & 1.14 & \textbf{110.1} & \textbf{9.66} & \textbf{1.15} & \textbf{1.14} \\
      \bottomrule
    \end{tabular}
    \caption{Results of cropping $8 \times 8$ patch at the center of the image where forget samples are 20\% (on the left) and 70\% (on the right) of a particular class. Forget samples were poisoned with the '$+$' sign in the CIFAR-10 dataset. $\mathbb{D}_f$ and $\mathbb{D}_r$ account for the forget samples and retain samples respectively. FID scores are compute with respect to retrained model, hence $\downarrow$ is better. Overall, the results highlight that our approach effectively unlearns forget samples and is closer to the retrained model.}
    \label{tab:sample_CIFAR10_res}
\end{table*}

\subsection{Results and Discussions}

Fig. \ref{fig:attack_CIFAR10} and Table \ref{tab:CIFAR10_res} present the comparison of various unlearning algorithms on CIFAR-10 dataset, where the forget samples are embedded with the '$+$' at the center of the image. As shown in Fig. \ref{fig:attack_CIFAR10}, our model, like the retrained model, successfully avoids generating the '$+$' symbol in its outputs, indicating effective unlearning of the forget class. Notably, both our model and the retrained model retain their ability to generalize lines, colors, and patterns from the retain samples, preserving essential generative capabilities. To provide a clearer comparison between the retrained model and our approach, Table \ref{tab:CIFAR10_res} presents the FID scores between the generated outputs of the retrained model and those of our method. A lower FID score indicates that our approach produces outputs that closely resemble those of the retrained model. At the same time, it achieves high-quality image generation, as indicated by a higher Inception Score (IS), reflecting crisp and detailed outputs.

When comparing results with VQ-GAN and diffusion models, we do not include a comparison with retrained model due to computational constraints. Instead, we compare our approach against the attack model and the original model. An effective unlearning process should increase the FID score relative to the attack model (indicating a shift away from poisoned patterns) while moving closer to the original model if the retained dataset provides sufficient patterns for reconstruction. In Table \ref{tab:VQ-GAN results}, we evaluate various unlearning algorithms by comparing their outputs against the attack model using $4 \times 4$ and $8 \times 8$ cropped patches on the ImageNet-1K dataset. Our approach consistently outperforms others in preserving the performance on retain samples. For forget samples, it generates outputs that are significantly distinct from those produced by the attack model, while maintaining high image quality (not necessarily accurate or reliable) and capturing similar semantics. Fig. \ref{fig:VQGAN_comp} further illustrates this comparison. As shown in the figure, our approach preserves the performance on the retain set while other approaches have traces of poisoned forget samples. For forget samples, our approach effectively removes the embedded '+' symbol, replacing it with patterns derived from the retain set, which may introduce some inaccuracies but no longer contain the memorized backdoor.

Furthermore, to assess whether our approach effectively prevents the generation of the embedded '$+$' symbol, we benchmark various baselines and state-of-the-art (SOTA) algorithms on diffusion models using the Places365 dataset.  Table \ref{tab:diff_model results} compares outputs with those of the \textit{original model} that was never trained on forget samples containing the '$+$' marker. Our model performs well on retained samples in most cases. For forget samples, a low FID score relative to the original model indicates that our approach effectively removes the '$+$' sign, as further evidenced by high IS and CLIP scores. This demonstrates that our model not only maintains high-quality output, but also ensures effective unlearning of sensitive data. 

One might argue from Fig. \ref{fig:attack_CIFAR10} and Fig. \ref{fig:VQGAN_comp} that our approach can reproduce images similar to the original ones. However, we demonstrate that this effect arises due to the presence of generic patterns within the cropped patch. Fig. \ref{fig:diff_masked_res} shows the generated outputs of our model for $4\times4$, $8\times8$, and $12\times12$ patches cropped at the center of the images. The figure clearly shows that as the patch size increases, our model produces accurate reconstructions for retained samples. In contrast, for forgotten samples, the model generates high-quality yet inaccurate images, reinforcing the effectiveness of our unlearning approach.

We compare the performance of all the approaches for unseen data in Table \ref{tab:OOD_model results}. We randomly sample the next 50 classes from ImageNet-1K and Places365 dataset for VQ-GAN and diffusion model respectively. It is clear from the results that our model has the best generative results on unseen data, particularly with diffusion models. We also perform T-SNE analysis to further validate the effectiveness of our approach, we randomly choose 50 outputs from retain and forget samples. We then compute the CLIP embedding vector for the generated out and the attack model. Fig. \ref{fig:tsne_combined} shows that the embedding vector from retain samples closely matches the ground truth, while the embedding vector from forget samples diverges. We also highlight the increasing distance ($\lambda$) with gradient ascent for CIFAR10 dataset in Fig. \ref{fig:model_dist}.

\subsection{Sample unlearning}

In order to evaluate the effectiveness of our approach on sample unlearning, we conduct experiments on CIFAR10 dataset. Specifically, we select two classes and unlearn 20\%, 30\%, 40\%, 50\%, and 75\% of their samples to analyze the impact of unlearning at different scales. We introduce a '$+$' marker into the forget samples before training and evaluate whether the model retains traces of this embedded pattern after unlearning. Fig. \ref{fig:sample_unl} shows the results of unlearning for varying size of forget samples. It is evident from the figure that our approach produces outputs most similar to those of the retrained model. With smaller size of forget samples, the rest of the approaches struggle to fully remove the embedded pattern, leading to partial regeneration of the '$+$' symbol. With increasing size of the forget samples, these methods exhibit even more pronounced retention of the poisoned marker.  In contrast, our method effectively removes the embedded backdoor, ensuring that the forget samples do not retain any visible traces of poisoning, regardless of the size of the forget samples.

Furthermore, Table \ref{tab:sample_CIFAR10_res} compares various unlearning algorithms on CIFAR-10 for 20\% and 75\% sample unlearning. The results demonstrate that, irrespective of the size of the forget set, our approach consistently generates high-quality images for both the retain and forget sets. With the comparison of FID scores, we show that our approach generates outputs that closely resemble those of retrained model. 

\section{Conclusion and Future work} \label{sec:conclusion}

In this work, we addressed the limitations of existing machine unlearning approaches for Image-to-Image (I2I) generative models. We challenged the current practice of treating unlearning as merely minimizing the distance between model outputs and Gaussian noise, arguing that this approach fails to account for a model’s ability to generalize patterns, even on forget samples. To overcome this limitation, we introduced a novel framework that ensures forget samples are treated as out-of-distribution (OOD) by leveraging gradient ascent to decouple the model’s parameters on these samples. We find that, after $T$ iterations of gradient ascent (where $T$ is application dependent), forget samples are effectively treated as OOD by the model. We establish formal $(\epsilon, \delta)$-unlearning guarantees with the gradient ascent for the first time. Following decoupling step, we fine-tune the model on the retained samples to preserve its performance. We further proposed an attack model to rigorously validate the effectiveness of our unlearning method. Empirical evaluations on large-scale datasets such as ImageNet-1K and Places365 demonstrated the superiority of our approach in achieving effective unlearning. Additionally, comparisons on the CIFAR-10 dataset using an AutoEncoder baseline showed that our method performs on par with a fully retrained model. We also show that our approach works well for sample unlearning, it effective unlearns the embedded backdoor regardless of the size of the forget samples.  

In this paper, we have utilized data poisoning attacks as an effective measure to audit the unlearning process. However, additional attack models, such as membership inference attacks \cite{duan2023diffusion} and data reconstruction attacks \cite{li2024gan}, also present promising avenues for auditing the effectiveness of unlearning. In future work, we plan to establish a benchmark to determine which attack models are most effective for auditing unlearning in I2I generative models. Additionally, we intend to explore how machine unlearning can be applied to address copyright infringement issues, providing a more robust framework for protecting intellectual property in generative models.

\backmatter

\bmhead{Acknowledgements}

This work was partially supported by the Wallenberg Al, Autonomous Systems and Software Program (WASP) funded by the Knut and Alice Wallenberg Foundation. The computations were enabled by the supercomputing resource Berzelius provided by National Supercomputer Centre at Linköping University and the Knut and Alice Wallenberg foundation.

\bibliography{bibliography}

\begin{appendices}

\section{Further Experiments} \label{sec:exp_normal_data}

\begin{table*}[h]
    \centering
    \begin{adjustbox}{width=1.2\textwidth,center}
    \begin{tabular}{ccccccc||cccccc}
      \toprule
      \multirow{3}{*}{Approach} & \multicolumn{6}{c||}{$4\times4$} & \multicolumn{6}{c}{$8\times8$} \\
      & \multicolumn{2}{c}{FID} & \multicolumn{2}{c}{IS} & \multicolumn{2}{c||}{CLIP} & \multicolumn{2}{c}{FID} & \multicolumn{2}{c}{IS} & \multicolumn{2}{c}{CLIP} \\ \cmidrule{2-13}
      & $\mathbb{D}_f$ & $\mathbb{D}_r$ & $\mathbb{D}_f$ & $\mathbb{D}_r$ & $\mathbb{D}_f$ & $\mathbb{D}_r$ & $\mathbb{D}_f$ & $\mathbb{D}_r$ & $\mathbb{D}_f$ & $\mathbb{D}_r$ & $\mathbb{D}_f$ & $\mathbb{D}_r$ \\
      \hline
      Max loss & 55.3 & 9.09 & 12.49 & 14.92 & 0.64 & 0.83 & 138.8 & 16.06 & 8.77 & 17.1 & 0.734 & 0.482 \\
      Random label & 21.23 & 8.81 & 13.9 & 15.18 & 0.80 & 0.84 & 78.78 & 14.81 & 11.27 & 17.62 & 0.74 & 0.64 \\
      Random encoder & 7.79 & 8.53 & 13.85 & 15.18 & 0.85 & 0.84 & 14.52 & 14.25 & 20.41 & 19.18 & 0.75 & 0.77 \\
      I2I SOTA & 21.92 & 8.61 & 13.95 & 15.24 & 0.79 & 0.84 & 82.90 & 14.89 & 11.75 & 18.75 & 0.74 & 0.60 \\
      Ours & 7.68 & 8.39 & 13.91 & 15.17 & 0.85 & 0.84 & 14.49 & 14.25 & 20.52 & 19.35 & 0.74 & 0.77 \\
      \bottomrule
    \end{tabular}
    \end{adjustbox}
    \caption{Comparison of various unlearning approaches with different cropped patches ($4\times4 \text{ and } 8\times8$) for VQ-GAN $\mathbb{D}_f$ and $\mathbb{D}_r$ account for the forget samples and retain samples, respectively. FID scores are computed with respect to original model. IS score highlight that our approach create good quality images even when the FID distance is significantly far from the attack model. Similarly, we find high CLIP values for our approach indicating that generated image still captures the semantics with an image (not just random noise).}
    \label{tab:VQ-GAN_results_original}
\end{table*}

\begin{table*}[h]
    \centering
    \begin{adjustbox}{width=1.2\textwidth,center}
    \begin{tabular}{ccccccc||cccccc}
      \toprule
      \multirow{3}{*}{Approach} & \multicolumn{6}{c||}{$4\times4$} & \multicolumn{6}{c}{$8\times8$} \\
      & \multicolumn{2}{c}{FID} & \multicolumn{2}{c}{IS} & \multicolumn{2}{c||}{CLIP} & \multicolumn{2}{c}{FID} & \multicolumn{2}{c}{IS} & \multicolumn{2}{c}{CLIP} \\ \cmidrule{2-13}
      & $\mathbb{D}_f$ & $\mathbb{D}_r$ & $\mathbb{D}_f$ & $\mathbb{D}_r$ & $\mathbb{D}_f$ & $\mathbb{D}_r$ & $\mathbb{D}_f$ & $\mathbb{D}_r$ & $\mathbb{D}_f$ & $\mathbb{D}_r$ & $\mathbb{D}_f$ & $\mathbb{D}_r$ \\
      \hline
      Max loss & 16.8 & 18.85 & 56.44 & 29.8 & 0.93 & 0.86 & 55.44 & 44.03 & 32.37 & 10.38 & 0.75 & 0.75 \\
      Random label & 16.63 & 11.68 & 55.55 & 34.13 & 0.93 & 0.90 & 51.50 & 33.67 & 55.60 & 23.45 & 0.93 & 0.24 \\
      Random encoder & 11.56 & 31.02 & 57.84 & 32.46 & 0.93 & 0.80 & 36.90 & 24.93 & 45.35 & 37.43 & 0.80 & 0.78 \\
      I2I SOTA & 47.93 & 18.14 & 44.27 & 27.05 & 0.67 & 0.18 & 113.71 & 17.79 & 11.24 & 30.74 & 0.67 & 0.84 \\
      Ours & 4.68 & 7.27 & 62.12 & 34.65 & 0.97 & 0.93 & 14.19 & 9.94 & 54.7 & 26.89 & 0.89 & 0.90 \\
      \bottomrule
    \end{tabular}
    \end{adjustbox}
    \caption{Comparison of various unlearning approaches for diffusion model with the output of \textit{the original model} for different cropped patches ($4 \times 4 \text{ and } 8\times8$). $\mathbb{D}_f$ and $\mathbb{D}_r$ account for the forget samples and retain samples, respectively. FID scores are computed with respect to original model. IS score highlight that our approach create good quality images even when the FID distance is significantly far from the attack model. Similarly, we find high CLIP values for our approach indicating that generated image still captures the semantics with an image (not just random noise).}
    \label{tab:diff_model_orig_results}
\end{table*}

In this section, we present additional results for the CIFAR10, Places365, and ImageNet-1K datasets. Table \ref{tab:VQ-GAN_results_original} and \ref{tab:diff_model_orig_results} compare several baseline methods and state-of-the-art approaches (referred to as the I2I SOTA approach) on $4\times4$ and $8\times8$ patches for the VQ-GAN and diffusion models, respectively. Interestingly, for smaller patches ($4\times4$), our approach achieves lower FID scores, indicating that the patterns learned from the retained samples are sufficient to effectively reconstruct smaller patches. However, this observation does not hold for larger patches ($8\times8$).

\begin{figure}
  \centering
    \includegraphics[width=\linewidth]{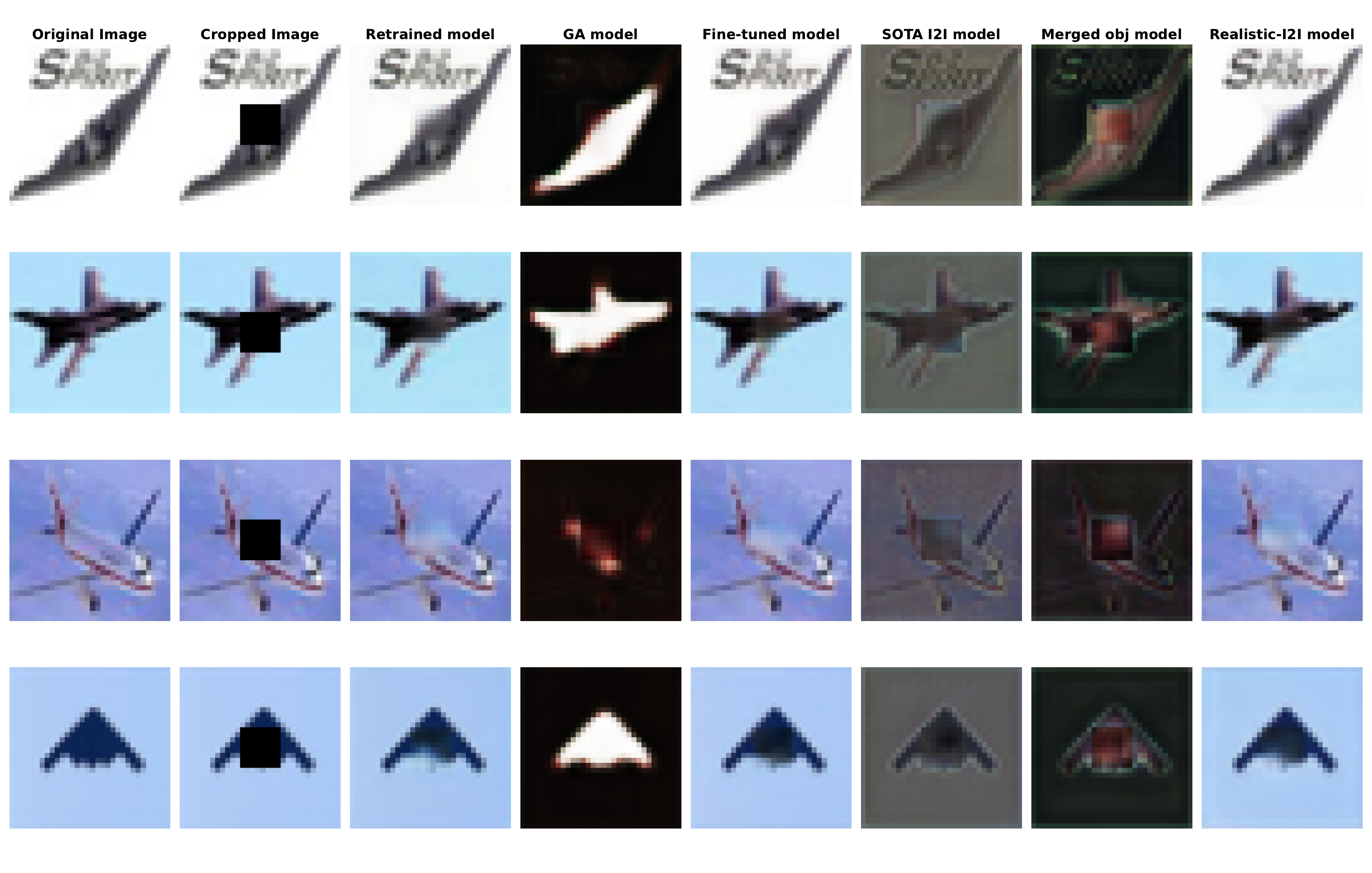}
    \caption{Comparison of various unlearning approaches, along with the retrained model for an AutoEncoder. The results clearly demonstrate that our method  produce outputs that are most similar to those of the retrained model.}
    \label{fig:res_CIFAR10_orig}
\end{figure}

Table \ref{tab:VQ-GAN_results_outpaint4x4original} and \ref{tab:VQ-GAN_results_outpaint8x8original} shows the results for image-outpainting. Here, we well our approach is far enough (this can be increased with even more gradient ascent steps). Fig. \ref{fig:res_CIFAR10_orig} and Table \ref{tab:CIFAR10_orig_res} shows the comparison of various baselines on CIFAR10 dataset. It is clear here that our approach has the most similar results to the retrained model from scratch.

\begin{table*}[h]
    \centering
    \begin{adjustbox}{width=1.2\textwidth,center}
    \begin{tabular}{ccccccc||cccccc}
      \toprule
      \multirow{3}{*}{Approach} & \multicolumn{6}{c||}{Original data} & \multicolumn{6}{c}{Poisoned data} \\
      & \multicolumn{2}{c}{FID} & \multicolumn{2}{c}{IS} & \multicolumn{2}{c||}{CLIP} & \multicolumn{2}{c}{FID} & \multicolumn{2}{c}{IS} & \multicolumn{2}{c}{CLIP} \\ \cmidrule{2-13}
      & $\mathbb{D}_f$ & $\mathbb{D}_r$ & $\mathbb{D}_f$ & $\mathbb{D}_r$ & $\mathbb{D}_f$ & $\mathbb{D}_r$ & $\mathbb{D}_f$ & $\mathbb{D}_r$ & $\mathbb{D}_f$ & $\mathbb{D}_r$ & $\mathbb{D}_f$ & $\mathbb{D}_r$ \\
      \hline
      Max loss & 103.6 & 12.05 & 13.71 & 15.63 & 0.63 & 0.74 & 181.04 & 14.44 & 10.14 & 15.79 & 0.66 & 0.74 \\
      Random label & 23.1 & 10.74 & 12.56 & 15.70 & 0.71 & 0.74 & 27.15 & 10.60 & 12.59 & 15.58 & 0.71 & 0.74 \\
      Random encoder & 8.29 & 10.97 & 13.47 & 16.12 & 0.78 & 0.74 & 22.76 & 10.6 & 12.97 & 15.50 & 0.78 & 0.74 \\
      I2I SOTA & 17.60 & 10.87 & 13.48 & 16.41 & 0.69 & 0.74 & 28.96 & 10.75 & 13.3 & 15.68 & 0.69 & 0.74 \\
      Ours & 8.36 & 11.15 & 13.29 & 16.25 & 0.78 & 0.74 & 22.70 & 11.03 & 13.18 & 16.19 & 0.78 & 0.74 \\
      \bottomrule
    \end{tabular}
    \end{adjustbox}
    \caption{Comparison of various unlearning approaches for image-outpainting on $4\times4$ cropped patches for VQ-GAN and VQ-GAN attack model (the model trained with '$+$ sign on forget samples). $\mathbb{D}_f$ and $\mathbb{D}_r$ account for the forget samples and retain samples, respectively. FID scores are computed with respect to original model. IS score highlight that our approach create good quality images even when the FID distance is significantly far from the attack model. Similarly, we find high CLIP values for our approach indicating that generated image still captures the semantics with an image (not just random noise).}
    \label{tab:VQ-GAN_results_outpaint4x4original}
\end{table*}

\begin{table*}[h]
    \centering
    \begin{adjustbox}{width=1.2\textwidth,center}
    \begin{tabular}{ccccccc||cccccc}
      \toprule
      \multirow{3}{*}{Approach} & \multicolumn{6}{c||}{Original data} & \multicolumn{6}{c}{Poisoned data} \\
      & \multicolumn{2}{c}{FID} & \multicolumn{2}{c}{IS} & \multicolumn{2}{c||}{CLIP} & \multicolumn{2}{c}{FID} & \multicolumn{2}{c}{IS} & \multicolumn{2}{c}{CLIP} \\ \cmidrule{2-13}
      & $\mathbb{D}_f$ & $\mathbb{D}_r$ & $\mathbb{D}_f$ & $\mathbb{D}_r$ & $\mathbb{D}_f$ & $\mathbb{D}_r$ & $\mathbb{D}_f$ & $\mathbb{D}_r$ & $\mathbb{D}_f$ & $\mathbb{D}_r$ & $\mathbb{D}_f$ & $\mathbb{D}_r$ \\
      \hline
      Max loss & 200.5 & 32.77 & 14.03 & 20.86 & 0.44 & 0.53 & 253.6 & 35.73 & 7.32 & 21.1 & 0.45 & 0.53 \\
      Random label & 89.22 & 27.13 & 12.75 & 20.68 & 0.50 & 0.54 & 53.51 & 25.71 & 12.92 & 21.04 & 0.50 & 0.54\\
      Random encoder & 20.05 & 28.0 & 21.11 & 25.34 & 0.63 & 0.53 & 24.13 & 28.66 & 18.24 & 21.37 & 0.62 & 0.52 \\
      I2I SOTA & 73.55 & 28.95 & 16.16 & 23.37 & 0.51 & 0.53 & 60.5 & 28.99 & 15.61 & 20.85 & 0.52 & 0.53 \\
      Ours & 19.93 & 27.61 & 21.89 & 26.71 & 0.63 & 0.53 & 23.15 & 29.28 & 20.05 & 23.70 & 0.62 & 0.51 \\
      \bottomrule
    \end{tabular}
    \end{adjustbox}
    \caption{Comparison of various unlearning approaches  for image-outpainting on $8\times8 $ cropped patches for VQ-GAN and VQ-GAN attack model (the model trained with '$+$ sign on forget samples). $\mathbb{D}_f$ and $\mathbb{D}_r$ account for the forget samples and retain samples, respectively. FID scores are computed with respect to original model. IS score highlight that our approach create good quality images even when the FID distance is significantly far from the attack model. Similarly, we find high CLIP values for our approach indicating that generated image still captures the semantics with an image (not just random noise).}
    \label{tab:VQ-GAN_results_outpaint8x8original}
\end{table*}
\begin{table*}[h]
    \centering
    \begin{tabular}{ccccc}
      \toprule
      \multirow{2}{*}{Approach} & \multicolumn{2}{c}{FID} & \multicolumn{2}{c}{IS} \\ \cmidrule{2-5}
      & $\mathcal{D}_f$ & $\mathcal{D}_r$ & $\mathcal{D}_f$ & $\mathcal{D}_r$ \\
      \hline
      GA model & 209.3 & 175.7 & 1.06 & 1.06 \\
      Fine-tuned model & 12.6 & 3.81 & 1.11 & 1.14 \\
      SOTA I2I model & 103.3 & 63.4 & 1.11 & 1.23 \\
      Merged obj model & 191.2 & 98.72 & 1.08 & 1.09 \\
      Realistic-I2I model & 12.1 & 4.77 & 1.11 & 1.14 \\
      \bottomrule
    \end{tabular}
    \caption{Results of cropping $8 \times 8$ patch at the center of the image in the CIFAR-10 dataset. $\mathbb{D}_f$ and $\mathbb{D}_r$ account for the forget samples and retain samples respectively. FID scores are compute with respect to retrained model, hence $\downarrow$ is better. Overall, the results highlight that our approach effectively unlearns forget samples and is closer to the retrained model.}
    \label{tab:CIFAR10_orig_res}
\end{table*}

\begin{figure*}
    \centering
    \includegraphics[width = 0.70\linewidth]{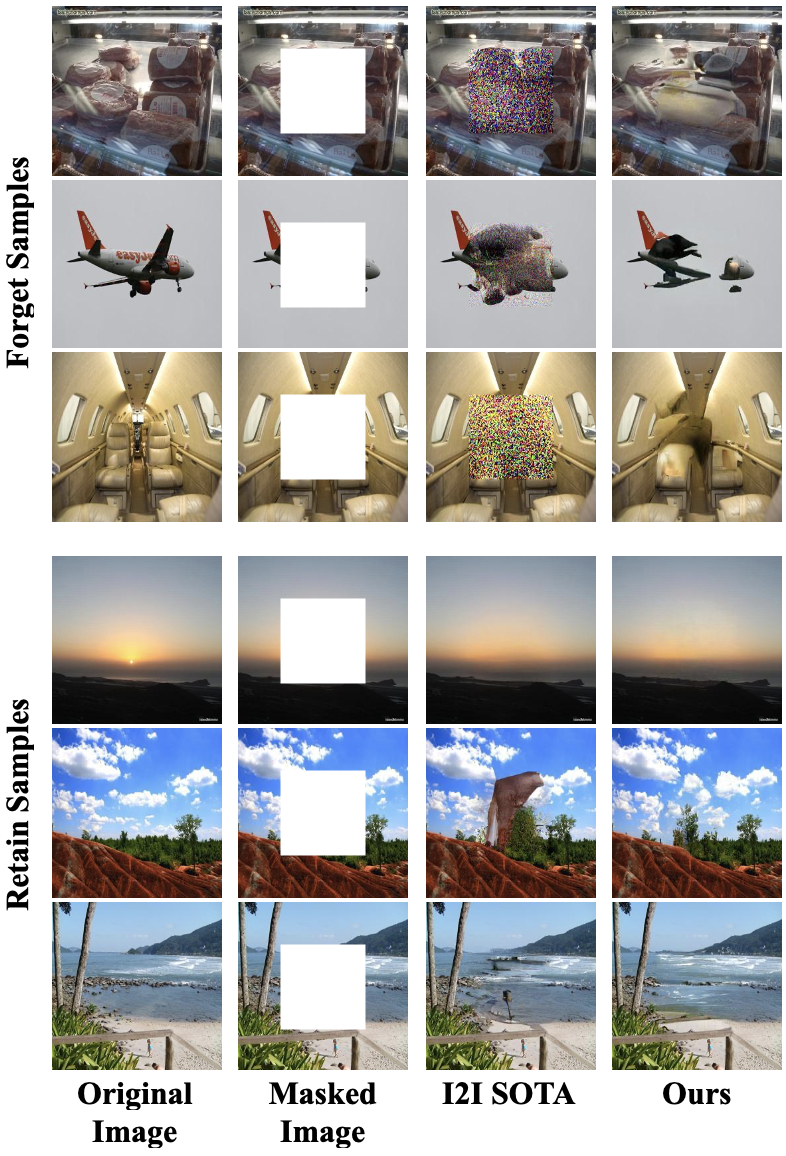}
    \caption{Results of cropping $8 \times 8$ patch at the center of the image on diffusion model.  The results demonstrate that our model effectively unlearns by producing incorrect/unreliable results instead of replacing it with Gaussian noise. For retain samples, our method does not reduces the quality of the images in contrast to the benchmark methods I2I SOTA. This happens due to the fine-tuning on the Gaussian noise. }
    \label{fig:VQGAN_comp_apn}
\end{figure*}

\end{appendices}
\end{document}